\documentclass[twoside,11pt]{article}

\usepackage{paralist,amsmath, amssymb}
\usepackage{booktabs}
\usepackage{multirow}
\usepackage{algorithm,algorithmic}
\usepackage{jmlr2e}

\usepackage[colorlinks,
            linkcolor=blue,
            citecolor=blue,
            urlcolor=magenta,
            linktocpage,
            plainpages=false]{hyperref}

\makeatletter
\newcounter{ALC@tempcntr}
\makeatother

\makeatletter
\AtBeginDocument{\Hy@breaklinkstrue}
\makeatother

\newtheorem{thm}{Theorem}

\newtheorem{cor}[thm]{Corollary}
 \newtheorem{ass}{Assumption}

\def \E {\mathrm{E}}
\def \x {\mathbf{x}}

\def \D {\mathbb{D}}

\def \H {\mathcal{H}}
\def \w {\mathbf{w}}
\def \R {\mathbb{R}}
\def \RR {\mathcal{R}}

\def \W {\mathcal{W}}
\def \N {\mathcal{N}}

\def \B {\mathcal{B}}

\def \Fh {\widehat{F}}
\def \Hh {\widehat{H}}

\def \wt {\widetilde{\w}}

\def \X {\mathcal{X}}

\def \K {\mathcal{K}}
\def \P {\mathbb{P}}

\def \wh {\widehat{\w}}

\def \B {\mathcal{B}}
\def \O {\widetilde{O}}
\def \OMG {\widetilde{\Omega}}

\DeclareMathOperator*{\argmin}{argmin}

\DeclareMathOperator*{\VC}{VC}

\def\abovestrut#1{\rule[0in]{0in}{#1}\ignorespaces}
\def\belowstrut#1{\rule[-#1]{0in}{#1}\ignorespaces}

\def\abovespace{\abovestrut{0.20in}}

\def\belowspace{\belowstrut{0.10in}}

\begin{document}

\title{Empirical Risk Minimization for Stochastic Convex Optimization: $O(1/n)$- and $O(1/n^2)$-type of Risk Bounds}

\author{\name Lijun Zhang \email zhanglj@lamda.nju.edu.cn\\
       \addr National Key Laboratory for Novel Software Technology\\
       Nanjing University, Nanjing 210023, China
       \AND
       \name Tianbao Yang \email tianbao-yang@uiowa.edu\\
       \addr Department of Computer Science\\
       the University of Iowa, Iowa City, IA 52242, USA
       \AND
       \name Rong Jin \email rongjin@cse.msu.edu\\
       \addr Alibaba Group, Seattle, USA}
\editor{}

\maketitle

\begin{abstract}
Although there exist plentiful theories of empirical risk minimization (ERM) for supervised learning, current theoretical understandings of ERM for a related problem---stochastic convex optimization (SCO), are limited. In this work, we strengthen the realm of ERM for SCO by  exploiting  smoothness and  strong convexity conditions to improve the risk bounds. First, we establish an $\O(d/n + \sqrt{F_*/n})$ risk bound when the random function is nonnegative, convex and smooth, and the expected function is Lipschitz continuous, where $d$ is the dimensionality of the problem, $n$ is the number of samples, and $F_*$ is the minimal risk. Thus, when $F_*$ is small we obtain an $\O(d/n)$ risk bound, which is analogous to the $\O(1/n)$ optimistic rate of ERM for supervised learning. Second, if the objective function is also $\lambda$-strongly convex, we prove an  $\O(d/n  + \kappa F_*/n )$ risk bound where $\kappa$ is the condition number, and improve it to $O(1/[\lambda n^2] + \kappa F_*/n)$ when $n=\OMG(\kappa d)$. As a result, we obtain an $O(\kappa/n^2)$ risk bound under the condition that $n$ is large and $F_*$ is small, which to the best of our knowledge, is the \emph{first} $O(1/n^2)$-type of risk bound of ERM. Third, we stress that the above results are established in a unified framework, which allows us to derive new risk bounds under weaker conditions, e.g., without convexity of the random function and  Lipschitz continuity of the expected function.  Finally, we demonstrate that to achieve an $O(1/[\lambda n^2] + \kappa F_*/n)$ risk bound for supervised learning,  the $\OMG(\kappa d)$ requirement on $n$ can be replaced with $\Omega(\kappa^2)$, which is dimensionality-independent.
\end{abstract}

\begin{keywords}
Empirical Risk Minimization, Stochastic Convex Optimization, Excess Risk
\end{keywords}

\section{Introduction}
Stochastic optimization occurs in almost all areas of science and engineering, such as  machine learning, statistics and operations research  \citep{Shapiro:2014:LSP}. In this problem, the goal is to optimize the value of an
expected objective function $F(\cdot)$ over some set $\W$, i.e.,
\begin{equation} \label{eqn:so}
\min_{\w \in \W} \  F(\w)=\E_{f\sim\P}\left[f(\w)\right],
\end{equation}
where $f(\cdot): \W \mapsto \R$ is a random function sampled from a (possibly unknown) distribution $\P$. A well-known special case is the risk minimization problem in supervised learning \citep{vapnik-1998-statistical,Nature:Statistical}, which takes the following form
\begin{equation} \label{eqn:supervised}
\min_{h \in \H} \  F(h)=\E_{(\x,y) \sim \D}\left[\ell(h(\x), y)\right],
\end{equation}
where $\H=\{h: \X \mapsto \R\}$ is a hypothesis class,  $(\x,y) \in \X \times \R$ is an instance-label pair sampled from a distribution $\D$, and  $\ell(\cdot,\cdot):\R \times \R \mapsto \R$ is certain loss. In this paper, we mainly focus on the convex version of (\ref{eqn:so}), namely stochastic convex optimization (SCO), where both the domain $\W$ and the expected function $F(\cdot)$ are convex.

Two classical approaches for solving stochastic optimization are stochastic approximation (SA) \citep{SA:Springer} and the sample average approximation (SAA), the latter of which is also referred to as empirical risk minimization (ERM) in the machine learning community \citep{vapnik-1998-statistical}. While both SA and ERM have been extensively studied in recent years \citep{Rademacher_Gaussian,Local_RC,Oracle_inequality,nemirovski-2008-robust,NIPS2011_4316}, most theoretical guarantees of ERM are restricted to the supervised learning problem in (\ref{eqn:supervised}). As pointed out in a seminal work of \cite{COLT:Shalev:2009}, the success of ERM for supervised learning cannot be directly extended to stochastic optimization. Actually, \cite{COLT:Shalev:2009} have constructed an instance of SCO that is learnable by SA but cannot be solved by ERM. Literatures about ERM for stochastic optimization (including SCO) are quite limited, and we still lack a full understanding of the theory.

In ERM, we are given $n$ i.i.d.~functions $f_1,\ldots,f_n$ sampled from $\P$, and minimize an empirical objective function:
\begin{equation} \label{eqn:empirical}
\min_{\w \in \W} \  \Fh(\w) = \frac{1}{n}\sum_{i=1}^n f_i(\w).
\end{equation}
Let $\wh \in\argmin_{\w \in \W}\Fh(\w)$ be the empirical minimizer. The performance of ERM is measured in terms of the excess risk defined as
\[
F(\wh)-\min_{\w \in \W} \  F(\w).
\]
State-of-the-art risk bounds of ERM include: an  $\O(\sqrt{d/n})$  bound  when the random function $f(\cdot)$  is Lipschitz continuous,\footnote{We use the $\O$ and $\OMG$ notations to hide constant factors as well as polylogarithmic factors in $d$ and $n$.} where $d$ is the dimensionality of $\w$; an $O(1/\lambda n)$ bound when $f(\cdot)$ is $\lambda$-strongly convex \citep{COLT:Shalev:2009}; and an $\O(d/\eta n)$ bound when $f(\cdot)$ is $\eta$-exponentially concave ($\eta$-exp-concave) \citep{arXiv:1605.01288}. From  existing studies of ERM for supervised learning \citep{Smooth:Risk}, we know that smoothness can be utilized to boost the risk bound. Thus, it is natural to ask whether smoothness can also be exploited to improve the performance of ERM for SCO. This paper provides an affirmative answer to this question. Indeed, we propose a general approach for analyzing the excess risk bound of ERM, which brings several improved risk bounds and new risk bounds as well.

\begin{table*}[t]
\centering\caption{Summary of Excess Risk Bounds of ERM for SCO. All bounds hold with high probability except the one marked by $^*$, which holds in expectation. Abbreviations: bounded $\rightarrow$ b, convex $\rightarrow$ c,  generalized linear $\rightarrow$ gl,  Lipschitz continuous $\rightarrow$ Lip, nonnegative $\rightarrow$ nn, strongly convex $\rightarrow$ sc,  smooth $\rightarrow$ sm, $\eta$-exponentially concave $\rightarrow$ $\eta$-exp.}
\label{tab:results}
\begin{tabular}{@{}c@{}c@{}c@{}c@{}c|c@{}}
\toprule  & &$f(\cdot)$&$\Fh(\cdot)$&$F(\cdot)$&Risk Bounds\\
 \hline
 \multicolumn{2}{c}{\multirow{2}{*}{\cite{COLT:Shalev:2009}} \!\!\!\!\!\!\!\!}  &  Lip &- &-& $\O(\sqrt{\frac{d}{n}})$\abovespace \\
 & &Lip \& sc &- &-& $O(\frac{1}{\lambda n})^*$ \abovespace \belowspace\\
\hline
 \multicolumn{2}{c}{\cite{arXiv:1605.01288}} & $\eta$-exp \& Lip \& b     &-&-&$\O(\frac{d}{\eta n})$ \abovespace \belowspace \\
\hline
\multirow{8}{*}{This work \ }& Theorem \ref{thm:smooth}  &    nn \& c \& sm&- &Lip& $\O(\frac{d}{n} + \sqrt{\frac{F_*}{n}})$ \abovespace \belowspace \\ \cline{2-6}
 & \multirow{2}{*}{Theorem \ref{thm:smooth:convex}} &  \multirow{2}{*}{nn \&  c \& sm}  &\multirow{2}{*}{-} & \multirow{2}{*}{Lip \& sc} &$\O(\frac{d}{n} + \frac{\kappa F_*}{n})$ \abovespace \\
 & & & && $O(\frac{1}{\lambda n^2} + \frac{\kappa F_*}{n})$ when $n=\OMG(\kappa d)$ \abovespace \belowspace\\\cline{2-6}
 & \multirow{2}{*}{Theorem \ref{thm:smooth:convex:2}} &\multirow{2}{*}{nn \& sm } &\multirow{2}{*}{c}& \multirow{2}{*}{sc}& $\O(\frac{\kappa d}{ n}+\frac{\kappa F_*}{n})=\O(\frac{\kappa d}{ n})$ \abovespace \\
  &  & & & & $O(\frac{1}{\lambda n^2} + \frac{\kappa F_*}{n})$ when $n=\OMG(\kappa^2 d)$ \abovespace \belowspace\\\cline{2-6}
& Theorem \ref{thm:smooth:weak} &nn \& sm &c &c&$\O(\sqrt{\frac{d}{n}}+\sqrt{\frac{F_*}{n}})=\O(\sqrt{\frac{d}{n}})$ \abovespace \belowspace\\\cline{2-6}
 & \multirow{2}{*}{Theorem \ref{thm:sup:learn}} &\multirow{2}{*}{nn \& sm \& gl } &\multirow{2}{*}{c}& \multirow{2}{*}{sc}& $O(\frac{\kappa }{ n} + \frac{\kappa F_*}{n})=O(\frac{\kappa }{ n} )$ \abovespace \\
&  & & & & $O(\frac{1}{\lambda n^2} + \frac{\kappa F_*}{n})$ when $n=\Omega(\kappa^2)$ \abovespace \belowspace\\
\bottomrule
\end{tabular}
\end{table*}

To state our results,  we first introduce some notations. Let $F_*= \min_{\w \in \W} F(\w)$ be the minimal risk, $\lambda$ be the modulus of strong convexity of $F(\cdot)$ and $L$ be the modulus of smoothness of $f(\cdot)$. Denote by $\kappa = L/\lambda$ the condition number of the problem. Our and previous results of ERM for SCO are summarized in Table~\ref{tab:results}, where we make explicit the assumptions on the random function $f(\cdot)$, the empirical function $\Fh(\w)$ and the expected function $F(\cdot)$. For our results of ERM for SCO, we assume the domain is bounded, and the random function is nonnegative.  We highlight the significance of this work as follows:
\begin{compactitem}
\item When $f(\cdot)$ is both convex and smooth and $F(\cdot)$ is Lipschitz continuous, we establish an $\O(d/n + \sqrt{F_*/n})$ risk bound (c.f.~Theorem \ref{thm:smooth}). In the optimistic case that $F_*$ is small, i.e., $F_*=O(d^2/n)$, we obtain an $\O(d/n)$ risk bound,  which is analogous to the $\O(1/n)$ optimistic rate of ERM for supervised learning \citep{Smooth:Risk} and also matches a recent lower bound of ERM for SCO~\citep[Theorem 3.10]{arXiv:1608.04414}.
\item If $F(\cdot)$ is also $\lambda$-strongly convex, we prove an $\O\left(d/n  + \kappa F_*/n \right)$ risk bound, and  improve it to $O(1/[\lambda n^2] + \kappa F_*/n)$ when $n=\OMG(\kappa d)$ (c.f.~Theorem~\ref{thm:smooth:convex}). Thus, if $n$ is large and $F_*$ is small, i.e., $F_*=O(1/n)$, we get an $O(\kappa/n^2)$ risk bound, which to the best of our knowledge, is the first $O(1/n^2)$-type of risk bound of ERM.
\item When  neither convexity is present in $f(\cdot)$ nor Lipschitz continuity is present in $F(\cdot)$, as long as $f(\cdot)$ is smooth, $\Fh(\cdot)$ is convex and $F(\cdot)$ is strongly convex, we still obtain an improved risk bound of  $O(1/[\lambda n^2] + \kappa F_*/n)$ when $n=\OMG(\kappa^2d)$, which will further implies an $O(\kappa /n^2)$ risk bound if $F_* = O(1/n)$ (c.f.~Theorem~\ref{thm:smooth:convex:2}).
\item If strong convexity is also absent in $F(\cdot)$, assuming $f(\cdot)$ is smooth and both $\Fh(\cdot)$ and $F(\cdot)$ are convex, we obtain an $\O(\sqrt{d/n})$ risk bound (c.f.~Theorem \ref{thm:smooth:weak}). This result breaks the barrier of non-learnability of bounded convex functions~\citep[Theorem 5.2]{arXiv:1608.04414} by exploiting the smoothness of random functions.
\item Finally, we extend the  $O(1/[\lambda n^2] + \kappa F_*/n)$ risk bound to supervised learning with a generalized linear form. Our  analysis shows that in this case, the lower bound of $n$ can be replaced with $\Omega(\kappa^2)$, which is dimensionality-independent (c.f.~Theorem \ref{thm:sup:learn}). Thus, this result can be applied to infinite dimensional cases, e.g., learning with kernels.
\end{compactitem}

\section{Related Work}
In this section, we give a brief introduction to previous work on stochastic optimization.
\subsection{ERM for Stochastic Optimization}
As we mentioned earlier, there are few works devoted to ERM for stochastic optimization. 
When $\W \subset \R^d$ is bounded and $f(\cdot)$  is Lipschitz continuous, \cite{COLT:Shalev:2009}  demonstrate that $\Fh(\w)$ converges to $F(\w)$ uniformly over $\W$ with an $\O(\sqrt{d/n})$ error bound that holds with high probability, implying an $\O(\sqrt{d/n})$ risk bound of ERM.  
They further establish an $O(1/\lambda n)$ risk bound of ERM that holds in expectation when $f(\cdot)$ is $\lambda$-strongly convex and Lipschitz continuous. Stochastic optimization
with exp-concave functions is studied recently \citep{NIPS2015_Exp},\footnote{Their excess risk bound is for a regularized empirical risk minimizer.}  and \cite{arXiv:1605.01288}  proves an $\O(d/\eta n)$ bound of ERM that holds with high probability when $f(\cdot)$ is  $\eta$-exp-concave, Lipschitz continuous, and bounded. Lower bounds of ERM for stochastic optimization is investigated by \cite{arXiv:1608.04414},  who exhibits (i) a lower bound of $\Omega(d/\epsilon^2)$ sample complexity  for uniform convergence that nearly matches the upper bound of \cite{COLT:Shalev:2009}; and (ii)  a lower bound of $\Omega(d/\epsilon)$ sample complexity of ERM, which is matched by our $\O(d/n + \sqrt{F_*/n})$ bound when $F_*$ is small.

It is worth mentioning the difference among  proof techniques in these  works. The uniform convergence result of \cite{COLT:Shalev:2009} leverages the covering number to bound $|\Fh(\w) - F(\w)|$ for any $\w\in\W$. The analysis for strongly convex functions by \cite{COLT:Shalev:2009} and exp-concave functions by \cite{NIPS2015_Exp} utilize the tool of stability, which only produces risk bounds that hold in expectation. A simple way to achieve a  high probability bound 
is to use ERM combined with a generic or specific boosting-the-confidence method~\citep{arXiv:1605.01288,DBLP:journals/iandc/HausslerKLW91}, but the guarantee is not directly on the empirical minimizer as noted by \cite{COLT:Shalev:2009}. The convergence of ERM given by \cite{arXiv:1605.01288} relies on a central condition or ``stochastic mixability''  of the exp-concave function. In this paper, we present a general approach for analyzing  ERM for SCO of smooth functions.  In particular, our analysis is based on a uniform convergence of $\nabla \Fh(\w) - \nabla \Fh(\w_*)$ to  $\nabla F(\w) - \nabla F(\w_*)$ for any $\w\in\W$,  and a concentration inequality of $\|\nabla \Fh(\w_*) - \nabla F(\w_*)\|$, where $\w_*$ is the optimal solution to  (\ref{eqn:so}).


\subsection{ERM for Supervised Learning} \label{sec:emr:supervised}
We note that there are extensive studies on ERM for supervised learning, and hence the review here is non-exhaustive.  In the context of supervised learning, the performance of ERM is closely related to the uniform convergence of $\Fh(\cdot)$ to $F(\cdot)$ over the hypothesis class $\H$ \citep{Oracle_inequality}. In fact, uniform convergence is a sufficient condition for learnability \citep{Understand:ML}, and in some special cases such as binary classification, it is also a necessary condition \citep{vapnik-1998-statistical}. The accuracy of uniform convergence, as well as the quality of the empirical minimizer, can be upper bounded in terms of the complexity of the hypothesis  class $\H$, including  data-independent measures such as the VC-dimension  and  data-dependent measures such as the Rademacher complexity.

Generally speaking, when $\H$ has finite VC-dimension, the excess risk can be upper bounded by $O(\sqrt{\VC(\H)/n})$, where $\VC(\H)$ is the VC-dimension of $\H$. If the loss $\ell(\cdot,\cdot)$ is Lipschitz continuous with respect to its first argument, we have a risk bound of $O(1/\sqrt{n} +  \RR_n(\H))$, where $\RR_n(\H)$ is the Rademacher complexity of $\H$. The Rademacher complexity typically scales as $\RR_n(\H)=O(1/\sqrt{n})$, e.g., $\H$ contains linear functions with low-norm, implying an $O(1/\sqrt{n})$ risk bound \citep{Rademacher_Gaussian}. There have been intensive efforts to derive rates faster than  $O(1/\sqrt{n})$ under various conditions \citep{Lee:1996:ICL,panchenko2002,Local_RC,Average:Stability}, such as low-noise \citep{Tsybakov04optimalaggregation},  smoothness \citep{Smooth:Risk}, strong convexity \citep{NIPS2008_3400}, to name a few amongst many. Specifically, when the random function $f(\cdot)$ is  nonnegative and smooth, \cite{Smooth:Risk} have established a risk bound of $\O(\RR_n^2(H)+\RR_n(H) \sqrt{F_*})$, reducing to an $\O(1/n)$ bound if $\RR_n(\H)=O(1/\sqrt{n})$ and $F_*=O(1/n)$. A  generalized linear form of (\ref{eqn:supervised}) is studied by \cite{NIPS2008_3400}, and a risk bound of $O(1/\lambda n)$ is proved if the expected function $F(\cdot)$ is $\lambda$-strongly convex.

\subsection{SA for  Stochastic Optimization}
Stochastic approximation (SA) solves the stochastic optimization problem via noisy observations of the expected function \citep{SA:Springer}. For brevity, we only discuss first-order methods for SCO, and in this case, $n$ is the number of stochastic gradients consumed by the algorithm. For Lipschitz continuous convex functions, stochastic gradient descent (SGD) exhibits the optimal $O(1/\sqrt{n})$ risk bound \citep{Problem_Complexity}. When the random function $f(\cdot)$ is nonnegative and smooth, SGD (with a suitable step size) has a risk bound of $O(1/n+\sqrt{F_*/n})$, becoming  $O(1/n)$  if $F_*=O(1/n)$ \citep[Corollary 4]{Smooth:Risk}. If $F(\cdot)$ is $\lambda$-strongly convex, some variants of SGD \citep{COLT:Hazan:2011,ICML2012Rakhlin} achieve an $O(1/\lambda n)$ rate which is known to be minimax optimal~\citep{IT:SCO}. For the square loss and the logistic loss, an $O(1/n)$ rate is attainable without any strong convexity assumptions \citep{NIPS2013_4900}. When the random function $f(\cdot)$ is $\eta$-exp-concave,  the online Newton step (ONS) is equipped with an $\O(d/ \eta n)$ risk bound \citep{ML:Hazan:2007,COLT:2015:Mahdavi}.

\section{Faster Rates of ERM}
We first introduce all the assumptions used in our analysis,  then present  theoretical results under different combinations of them, and finally discuss a special case of supervised learning.
\subsection{Assumptions}
In the following, we use $\|\cdot\|$ to denote the $\ell_2$-norm of vectors.
\begin{ass}\label{ass:1}
The domain $\W$ is a convex subset of $\R^d$, and is bounded by $R$, that is,
\begin{equation} \label{eqn:domain:R}
\|\w\| \leq R, \ \forall \w \in \W.
\end{equation}
\end{ass}
\begin{ass}\label{ass:2}
The random function $f(\cdot)$ is nonnegative, and $L$-smooth  over $\W$, that is,
\begin{equation} \label{eqn:f:smooth}
\left \|\nabla f(\w)-\nabla f(\w') \right\| \leq L \|\w-\w'\|, \ \forall \w, \w' \in \W, \ f \sim \P.
\end{equation}
\end{ass}
\begin{ass}\label{ass:3}
The expected function $F(\cdot)$ is $G$-Lipschitz continuous over $\W$, that is,
\begin{equation} \label{eqn:F:Lipschitz}
  |F(\w)-F(\w')| \leq G \|\w -\w'\|, \ \forall \w, \w' \in \W.
\end{equation}
\end{ass}
\begin{ass}\label{ass:4} We use different combinations of the following assumptions on convexity.
\begin{itemize}
\item [\bf (a)] The expected function $F(\cdot)$ is convex over $\W$.
\item [\bf (b)] The expected function $F(\cdot)$ is $\lambda$-strongly convex over $\W$, that is,
\begin{equation} \label{eqn:F:strong}
   F(\w)+\langle \nabla F(\w), \w' -\w \rangle + \frac{\lambda}{2} \|\w'-\w\|^2 \leq F(\w'), \ \forall \w, \w' \in \W.
\end{equation}
\item[\bf (c)] The empirical function  $\Fh(\cdot)$ is convex.
\item [\bf (d)] The random function  $f(\cdot)$ is convex.
\end{itemize}
\end{ass}
\paragraph{Remark 1} First, note that {\bf Assumption~\ref{ass:4}(a)} is implied by either {\bf Assumption~\ref{ass:4}(b)} or {\bf Assumption~\ref{ass:4}(d)}, and {\bf Assumption~\ref{ass:4}(c)} is implied by {\bf Assumption~\ref{ass:4}(d)}. Second, the smoothness assumption of $f(\cdot)$  implies the  expected function $F(\cdot)$ is  $L$-smooth. By Jensen's inequality, we have
\[
\left\|\nabla F(\w)-\nabla F(\w') \right\| \leq \E_{f \sim \P}\left\|\nabla f(\w)-\nabla f(\w') \right\|\leq L \|\w-\w'\|, \ \forall \w, \w' \in \W.
\]
Similarly, the  empirical function  $\Fh(\cdot)$ is also $L$-smooth. The \emph{condition number} $\kappa$ of $F(\cdot)$ is defined as the ratio between $L$ and $\lambda$, i.e., $\kappa=L/\lambda \geq 1$.

\subsection{Risk Bounds for SCO}
Let
$\w_* \in\argmin_{\w \in \W} F(\w) \textrm{ and } \wh \in\argmin_{\w \in \W}\Fh(\w)$
be optimal solutions to (\ref{eqn:so}) and (\ref{eqn:empirical}), respectively. We first present an excess risk bound under the smoothness condition.
\begin{thm} \label{thm:smooth}
For any $0< \delta < 1$, define
\begin{eqnarray}
M =   \sup \limits_{f \sim \P} \|\nabla f(\w_*)\|, \label{eqn:defin} \\
 C(\varepsilon)= 2  \left( \log \frac{2}{\delta} + d\log\frac{6 R}{\varepsilon}\right). \label{eqn:c:eps}
\end{eqnarray}
Under {\bf Assumptions~\ref{ass:1}}, {\bf\ref{ass:2}}, {\bf\ref{ass:3}}, and {\bf\ref{ass:4}(d)}, with probability at least $1 - 2\delta$, we have
\begin{equation}\label{eqn:smooth:r1}
\begin{split}
& F(\wh) - F(\w_*) \\
\leq &   \frac{16 R^2 LC(\varepsilon)}{n}+   \frac{8RM \log(2/\delta)}{n}     + 8R \sqrt{\frac{2 L F_* \log(2/\delta)}{n} }  + \left( 8 R L+ G  + \frac{4 R LC(\varepsilon) }{n} \right)\varepsilon,
\end{split}
\end{equation}
where $F_*=F(\w_*)$ is the minimal risk.
\end{thm}
By choosing $\varepsilon$ small enough, the last term in (\ref{eqn:smooth:r1}) that contains $\varepsilon$ becomes non-dominating. To be specific, we have the following corollary.
\begin{cor}By setting $\varepsilon=1/n$ in Theorem~\ref{thm:smooth}, we have
 $C( 1/n )= 2  \left( \log \frac{2}{\delta} + d\log(6 n R)\right)=\Theta(  d \log n)$,
and with  high probability
\[
 F(\wh) - F(\w_*)= O\left( \frac{d \log n}{n} + \sqrt{\frac{F_*}{n}} \right)=\O\left( \frac{d}{n} + \sqrt{\frac{F_*}{n}} \right).
\]
\end{cor}
\paragraph{Remark 2} The above corollary implies that under the smoothness and other common assumptions, ERM achieves  an $\O(d/n + \sqrt{F_*/n})$ risk bound for SCO. When the minimal risk is small, i.e., $F_*=O(d^2/n)$, the rate is improved to $\O(d/n)$. Note that even under the smoothness assumption, the linear dependence on $d$ is unavoidable~\citep[Theorem 3.7]{arXiv:1608.04414}.

We next present excess risk bounds under both the smoothness and strong convexity conditions.
\begin{thm} \label{thm:smooth:convex}
Under {\bf Assumptions~\ref{ass:1}}, {\bf\ref{ass:2}}, {\bf\ref{ass:3}}, {\bf\ref{ass:4}(b)}, and {\bf\ref{ass:4}(d)}, with probability at least $1 - 2\delta$, we have
\begin{equation}\label{eqn:main:1}
\begin{split}
& F(\wh) - F(\w_*) \\
\leq  & \frac{16 R^2 LC(\varepsilon) }{n} +    \frac{8RM\log(2/\delta)}{n}  + \frac{8 L F_* \log(2/\delta)}{\lambda n} + \left(8 R L+G    + \frac{4 R LC(\varepsilon) }{n} \right)\varepsilon.
\end{split}
\end{equation}
Furthermore, if
\begin{equation} \label{eqn:lower:n}
n \geq \frac{4 LC(\varepsilon)}{\lambda}=4 \kappa C(\varepsilon),
\end{equation}
 we also have
\begin{equation}\label{eqn:main:2}
 F(\wh) - F(\w_*) \leq  \frac{32M^2\log^2(2/\delta)}{\lambda n^2}+  \frac{128 L F_* \log(2/\delta)}{\lambda n} + \left( \frac{128 L^2 \varepsilon^2}{\lambda} +16 G \varepsilon +4 \lambda \varepsilon^2 \right).
\end{equation}
\end{thm}

The above theorem can be simplified by choosing different values of $\varepsilon$.
\begin{cor}\label{cor:strong}By setting $\varepsilon=1/n$ in Theorem~\ref{thm:smooth:convex}, we have $C(1/n)=\Theta(  d \log n)$, and with high probability
\[
 F(\wh) - F(\w_*)= O\left( \frac{d \log n}{n} + \frac{\kappa F_*}{ n} \right)=\O\left( \frac{d}{n} + \frac{\kappa F_*}{ n} \right).
\]
Setting $\varepsilon=1/n^2$, we have
$ C( 1/n^2 )= 2  \left( \log \frac{2}{\delta} + d\log(6 n^2 R)\right)=\Theta(  d \log n)$
and when
$n = \Omega( \kappa d \log n) = \OMG(\kappa d)$,
with high probability
\[
 F(\wh) - F(\w_*)= O\left( \frac{1}{\lambda n^2} + \frac{\kappa F_*}{ n} \right).
\]
\end{cor}
\paragraph{Remark 3} The first part of Corollary \ref{cor:strong} shows that ERM enjoys an $\O\left(d/n  + \kappa F_*/n \right)$ risk bound  for stochastic optimization of strongly convex and smooth functions.
In the literature, the most comparable result is the $O(1/\lambda n)$ risk bound proved by \cite{COLT:Shalev:2009} but with striking differences highlighted in Table~\ref{tab:results}. 
Since the risk bound of \cite{COLT:Shalev:2009} is independent of the dimensionality $d$, it is natural to ask whether it is possible to prove a dimensionality-independent $\O(\kappa/ n)$  bound that holds with high probability. The second part of  Corollary \ref{cor:strong} indeed provides such a bound, but under an additional condition $n =  \OMG(\kappa d)$.

\paragraph{Remark 4} The second part implies that when $n$ is large enough, i.e., $n =  \OMG(\kappa d)$, the risk bound can be tightened to $O(1/[\lambda n^2] + \kappa F_*/n)$. In particular, when the minimal risk is small, i.e., $F_*=O(1/n)$, we obtain an 
$O(\kappa/n^2)$ bound. To the best of our knowledge, this is the first $O(1/n^2)$-type of risk bound of ERM, and even in the studies of stochastic approximation, we have not found similar theoretical guarantees. Finally, it is worth to point out the following two features of the second part:
\begin{compactitem}
\item Although the lower bound of $n$ depends on $d$, the risk bound is independent of $d$.
\item The domain size $R$ only appears in the lower bound of $n$, and the dependence is logarithmic.
\end{compactitem}

Our next result shows that the individual convexity assumption, i.e., {\bf Assumption~\ref{ass:4}(d)}, and the Lipschitz continuity assumption, i.e., {\bf Assumptions~\ref{ass:3}}, in Theorem \ref{thm:smooth:convex} can be relaxed. To be specific, {\bf Assumptions~\ref{ass:4}(d)} and {\bf\ref{ass:3}} can be replaced with {\bf Assumption \ref{ass:4}(c)}.
\begin{thm} \label{thm:smooth:convex:2}
Under {\bf Assumptions~\ref{ass:1}}, {\bf \ref{ass:2}}, {\bf\ref{ass:4}(b)}, and {\bf\ref{ass:4}(c)}, with probability at least $1 - 2\delta$, we have
\begin{equation}\label{eqn:main:1:2}
\begin{split}
 F(\wh) - F(\w_*) \leq  &  \frac{4R^2LC(\varepsilon)}{n} + \frac{4 R^2 L^2 C(\varepsilon) }{\lambda n}  +  \frac{4 RM\log(2/\delta) }{n} + \frac{8 L F_* \log(2/\delta)}{\lambda n} \\
& +\left(4  R L  +2 R L  \sqrt{\frac{C(\varepsilon) }{n} } + \frac{2 R L C(\varepsilon) }{n}  \right) \varepsilon.
\end{split}
\end{equation}
Furthermore, if
\begin{equation} \label{eqn:lower:n:2}
n \geq \frac{25 L^2C(\varepsilon)}{\lambda^2} = 25 \kappa^2 C(\varepsilon),
\end{equation}
 we also have
\begin{equation}\label{eqn:main:2:2}
 F(\wh) - F(\w_*) \leq \frac{8M^2\log^2(2/\delta)}{\lambda n^2} + \frac{32 L F_* \log(2/\delta)}{\lambda n}  + \left(\frac{32 L^2}{\lambda}  + \frac{416 \lambda  }{625}\right)\varepsilon^2.
\end{equation}
\end{thm}
We have the following corollary to simplify the above theorem.
\begin{cor}\label{cor:strong:2} By setting $\varepsilon=1/n$ in Theorem~\ref{thm:smooth:convex:2}, we have $C(1/n)=\Theta(  d \log n)$, and with high probability
\[
 F(\wh) - F(\w_*)= O\left( \frac{ \kappa d \log n}{ n} + \frac{\kappa F_*}{ n} \right)=\O\left( \frac{\kappa d }{n}  + \frac{\kappa F_*}{ n}  \right)=\O\left( \frac{\kappa d }{n} \right).
\]
Setting $\varepsilon=1/n^2$, we have $C(1/n^2)=\Theta(d \log n)$, and when
$n = \Omega\left( \kappa^2 d \log n\right) = \OMG(\kappa^2 d)$,
with high probability
\[
 F(\wh) - F(\w_*)= O\left( \frac{1}{\lambda n^2} + \frac{\kappa F_*}{ n} \right).
\]
\end{cor}

\paragraph{Remark 5} Comparing the first part of Corollaries~\ref{cor:strong:2} and \ref{cor:strong}, we observe that without the  individual convexity and Lipschitz continuity, the risk bound is increased from $\O\left(d/n  + \kappa F_*/n \right)$ to $\O( \kappa d /n+ \kappa F_*/n)$.

\paragraph{Remark 6}  Comparing the second part of Corollaries~\ref{cor:strong:2} and \ref{cor:strong}, we can see that the risk bound is on the same order, but the lower bound of $n$ is increased by a factor of $\kappa$. It is interesting to mention that a similar phenomenon also happens in stochastic approximation. Recently, a variance reduction technique named SVRG  \citep{NIPS13:ASGD} or EMGD \citep{NIPS:13:Mixed} was proposed for stochastic optimization when both full gradients and stochastic gradients are available. In the analysis, SVRG assumes the stochastic function is convex, while EMGD does not. From their theoretical results, we observe that the individual convexity leads to a difference of $\kappa$ factor in the sample complexity of stochastic gradients.

Finally, we want to mention that even when the strong convexity assumption  in Theorem~\ref{thm:smooth:convex:2} is missing, a risk bound of $\O(\sqrt{d/n})$ is still attainable.
\begin{thm} \label{thm:smooth:weak}
Under {\bf Assumptions~\ref{ass:1}}, {\bf \ref{ass:2}}, {\bf\ref{ass:4}(a)},  and {\bf\ref{ass:4}(c)}, with probability at least $1 - 2\delta$, we have
\[
\begin{split}
 F(\wh) - F(\w_*) \leq   & \frac{4R^2LC(\varepsilon) }{n}+  \frac{4RM\log(2/\delta) }{n} + 4R^2 L \sqrt{\frac{C(\varepsilon) }{n} }   + 2R \sqrt{\frac{8 L F_* \log(2/\delta)}{n}} \\
& +\left( 4  R L  +2 R L  \sqrt{\frac{C(\varepsilon) }{n} } + \frac{2 R L C(\varepsilon) }{n} \right) \varepsilon.
\end{split}
\]
\end{thm}
\paragraph{Remark 7} In a recent work, \cite{arXiv:1608.04414} shows that SCO without the Lipschitz condition cannot be solved by ERM. Theorem \ref{thm:smooth:weak} exhibits that as long as the random function is smooth,  SCO is learnable by ERM.

\subsection{Risk Bounds for Supervised Learning}
If the conditions of Theorem \ref{thm:smooth:convex} or Theorem \ref{thm:smooth:convex:2} are satisfied, we can directly use them to establish an $O(1/[\lambda n^2] + \kappa F_*/n)$ risk bound for supervised learning. However, a major limitation of these theorems is that the lower bound of $n$ depends on the dimensionality $d$, and thus cannot be applied to infinite dimensional cases, e.g., kernel methods \citep{Learning_with_kernels}. In this section, we exploit the structure of supervised learning to make the theory dimensionality-independent.

We focus on the generalized linear form of supervised learning:
\begin{equation}\label{eqn:generalized}
\min_{\w \in \W}\  F(\w)=\E_{(\x,y) \sim \D}\left[\ell(\langle \w, \x\rangle, y)\right] + r(\w),
\end{equation}
where $\ell(\langle \w, \x\rangle, y)$ is the loss of predicting $\langle \w, \x\rangle$ when the true target is $y$, and $r(\cdot)$ is a regularizer. Given $n$ training examples $(\x_1,y_1),\ldots, (\x_n,y_n)$ independently sampled from $\D$, the empirical objective is
\[
\min_{\w \in \W} \  \Fh(\w) = \frac{1}{n}\sum_{i=1}^n \ell(\langle \w, \x_i\rangle, y_i) + r(\w).
\]
We define
\[
H(\w)=\E_{(\x,y) \sim \D}\left[\ell(\langle \w, \x\rangle, y)\right] \textrm{ and } \Hh(\w)=\frac{1}{n}\sum_{i=1}^n \ell(\langle \w, \x_i\rangle, y_i)
\]
to capture the stochastic component.

Besides {\bf\ref{ass:4}(b)} and {\bf\ref{ass:4}(c)}, we introduce the following additional assumptions. We abuse the same notation $\|\cdot\|$ to denote the norm  induced by the inner product of a Hilbert space.
\begin{ass}\label{ass:4-5}
The domain $\W$ is a convex subset of a Hilbert space $\H$, and is bounded by $R$, that is,
\begin{equation} \label{eqn:domain:R:New}
\|\w\| \leq R, \ \forall \w \in \W.
\end{equation}
\end{ass}
\begin{ass}\label{ass:5}
The norm of the random data $\x\in\H$ is upper bounded by a constant $D$, that is,
\begin{equation} \label{eqn:at}
  \|\x\|  \leq D, \ \forall (\x, y) \sim \D.
\end{equation}
\end{ass}
\begin{ass}\label{ass:6}
For any $(\x, y) \sim \D$, $\ell(\cdot, y)$ is nonnegative, and $\beta$-smooth over $[-DR,DR]$, that is,
  \begin{equation} \label{eqn:smooth:gt}
  |\ell'(u, y)-\ell'(v, y)| \leq \beta |u-v|, \ \forall u, v \in [-DR,DR].
  \end{equation}
\end{ass}
\begin{ass}\label{ass:7}
The regularizer $r(\cdot)$ is $P$-Lipschitz continuous over $\W$, that is,
\begin{equation} \label{eqn:r:Lipschitz}
  |r(\w)-r(\w')| \leq P \|\w -\w'\|, \ \forall \w, \w' \in \W.
\end{equation}
\end{ass}

\paragraph{Remark 8} The above assumptions allow us to model many popular losses in machine learning, such as (regularized) least squares and (regularized) logistic regression.  {\bf Assumptions~\ref{ass:5}} and {\bf\ref{ass:6}} imply the random function $\ell(\langle \cdot, \x\rangle, y)$ is $\beta D^2$-smooth over $\W$. To see this, for any $\w, \w' \in \W$, we have
\[
\begin{split}
&\left\|\nabla  \ell(\langle \w, \x\rangle, y) -\nabla \ell(\langle \w', \x\rangle, y) \right\|= \left\| \ell'(\langle \w, \x\rangle, y) \x- \ell'(\langle \w', \x\rangle, y) \x \right\| \\
\overset{\text{(\ref{eqn:at})}}{\leq} & D| \ell'(\langle \w, \x\rangle, y) - \ell'(\langle \w', \x\rangle, y)| \overset{\text{(\ref{eqn:smooth:gt})}}{\leq}  \beta D |\langle \w, \x\rangle-\langle \w', \x\rangle|\overset{\text{(\ref{eqn:at})}}{\leq}\beta D^2 \|\w-\w'\|.
\end{split}
\]
By Jensen's inequality, $H(\cdot)$ is also $\beta D^2$-smooth.  Notice that $\beta D^2$ is the modulus of smoothness of $H(\cdot)$, and $\lambda$ is the modulus of strong convexity of $F(\cdot)$. With a slight abuse of notation, we define $L=\beta D^2$, and the condition number $\kappa$ as the ratio between $L$ and $\lambda$, i.e., $\kappa=L/\lambda$. Finally, we note that the regularizer $r(\cdot)$ could be \emph{non-smooth}.

Recall that $\w_*  \in\argmin_{\w \in \W} F(\w)$ and $\wh \in\argmin_{\w \in \W}\Fh(\w)$.  We have the following excess risk bound of ERM for supervised learning.
\begin{thm} \label{thm:sup:learn}
For any $0< \delta < 1$, define
\begin{eqnarray}
M =  \sup \limits_{(\x,y) \sim \D} \|\nabla  \ell(\langle \w_*, \x\rangle, y) \|, \label{eqn:sup:M} \\
C=4 \left( 8 +  \sqrt{2 \log \frac{\lceil 2\log_2(n) + \log_2(2R)\rceil}{\delta}} \right),\label{eqn:defin:C}\\
H_*=H(\w_*) = F(\w_*)-r(\w_*).
\end{eqnarray}
Under {\bf Assumptions} {\bf\ref{ass:4}(b)},  {\bf\ref{ass:4}(c)},  {\bf\ref{ass:4-5}},  {\bf\ref{ass:5}}, {\bf\ref{ass:6}}, and {\bf\ref{ass:7}},  with probability at least $1 - 2\delta$, we have
\begin{equation}\label{eqn:sup:1}
F(\wh) - F(\w_*) \leq \max\left(\frac{M +P}{n^2}+ \frac{L}{2n^4},  \frac{4 R^2 L^2 C^2 }{\lambda n}+ \frac{4 RM\log(2/\delta)}{n}  +\frac{8 L H_* \log(2/\delta)}{\lambda n}\right).
\end{equation}
Furthermore, if
\begin{equation} \label{eqn:sup:lower:n}
n\geq \frac{16 L^2  C^2}{\lambda^2} =16 \kappa^2 C^2,
\end{equation}
with probability at least $1 - 2\delta$, we have
\begin{equation}\label{eqn:sup:2}
F(\wh) - F(\w_*) \leq \max\left(\frac{M +P}{n^2}+ \frac{L}{2n^4},  \frac{8M^2\log^2(2/\delta)}{\lambda n^2}+\frac{16 L H_* \log(2/\delta)}{\lambda n}\right).
\end{equation}
\end{thm}

\paragraph{Remark 9} The first part of Theorem~\ref{thm:sup:learn} presents an $O(\kappa/n)$ risk bound,\footnote{For brevity, we treat $C$ as a constant because it only has a \emph{double} logarithmic dependence on $n$.} similar to the $O(1/\lambda n)$ risk bound of \cite{NIPS2008_3400}. The second part is an $O(1/[\lambda n^2] + \kappa H_*/n)$ risk bound, and in this case, the lower bound of $n$ is $\Omega(\kappa^2)$,  which is dimensionality-independent. Thus, Theorem~\ref{thm:sup:learn} can be applied even when the dimensionality is infinite. Generally speaking, the regularizer $r(\cdot)$ is nonnegative, and thus $H_* \leq F_*$. So, the second bound is even better than those in Theorems \ref{thm:smooth:convex} and  \ref{thm:smooth:convex:2}. Finally, we note that Theorem~\ref{thm:sup:learn} should be treated as a counterpart of  Theorem~\ref{thm:smooth:convex:2} for supervised learning, because both of them do not rely on the individual complexity, i.e., {\bf Assumption~\ref{ass:4}(d)}. One may wonder whether it is possible to derive a counterpart of Theorem~ \ref{thm:smooth:convex}, that is, whether it is possible to utilize the individual convexity to reduce the lower bound of $n$ by a factor of $\kappa$. We will investigate this question as a future work.
\section{Analysis}
We here present the proofs of main theorems. The omitted ones can be found in appendices.
\subsection{The Key Idea}
 By the convexity of $\Fh(\cdot)$ and the optimality condition of $\wh$ \citep{Convex-Optimization}, we have
\begin{equation} \label{eqn:opt}
\langle \nabla \Fh(\wh) , \w - \wh \rangle \geq 0, \ \forall \w \in \W .
\end{equation}
Our theoretical analysis is built upon the following inequality:
\begin{equation} \label{eqn:key}
\begin{aligned}
& F(\wh) - F(\w_*) + \frac{\lambda}{2}\|\wh - \w_*\|^2 \leq  \langle \nabla F(\wh), \wh - \w_* \rangle\\
=& \langle \nabla F(\wh) - \nabla F(\w_*), \wh - \w_* \rangle + \langle \nabla F(\w_*), \wh - \w_*\rangle \\
=& \langle \nabla F(\wh) - \nabla F(\w_*) - [\nabla \Fh(\wh) - \nabla \Fh(\w_*)], \wh - \w_* \rangle \\
&+ \langle \nabla \Fh(\wh) - \nabla \Fh(\w_*) + \nabla F(\w_*), \wh - \w_* \rangle \\
\overset{\text{(\ref{eqn:opt})}}{\leq} & \langle \nabla F(\wh) - \nabla F(\w_*) - [\nabla \Fh(\wh) - \nabla \Fh(\w_*)], \wh - \w_* \rangle + \langle \nabla F(\w_*) - \nabla \Fh(\w_*), \wh - \w_*\rangle,
\end{aligned}
\end{equation}
where $\lambda>0$ is the strong convexity modulus of $F(\cdot)$ if exists otherwise it is zero.

In Theorems~\ref{thm:smooth}, \ref{thm:smooth:convex}, \ref{thm:smooth:convex:2}, and \ref{thm:smooth:weak}, we utilize the covering number to upper bound the first term on the last line of (\ref{eqn:key}), and thus introduce a linear dependence on the dimensionality $d$. In Theorem~\ref{thm:sup:learn}, we use the  Rademacher complexity to upper bound it, leading to a dimensionality-independent bound. The second term on the last line of (\ref{eqn:key}) is upper bounded by the concentration inequality for vectors, which produces a quantity containing $F_*$.

\subsection{Proof of Theorem~\ref{thm:smooth}}
We set $\lambda=0$ in (\ref{eqn:key}), and upper bound the last line as
\begin{equation} \label{eqn:main:inequality}
\begin{split}
& F(\wh) - F(\w_*) \\
\leq &\left( \underbrace{\left\| \nabla F(\wh) - \nabla F(\w_*) - [\nabla \Fh(\wh) - \nabla \Fh(\w_*)] \right\| }_{:=A_1}+ \underbrace{\left\| \nabla F(\w_*) - \nabla \Fh(\w_*) \right\|}_{:=A_2} \right)\left\| \wh - \w_* \right\|.
\end{split}
\end{equation}
We first bound $A_1$. Let $\N(\W , \varepsilon)$ be the $\varepsilon$-net of $\W$ with minimal cardinality, which is referred to as the covering numbers.\footnote{A subset $\N \subseteq \K$ is called an $\varepsilon$-net of $\K$ if for every $\w \in  \K$ one can find $\wt \in \N$ so that $\|\w-\wt\|\leq \varepsilon$.}  Based on the concentration inequality of vectors \citep{Smale:learning}, we  establish a uniform convergence of $\nabla F(\w) - \nabla F(\w_*)$ to $\nabla \Fh(\w) - \nabla \Fh(\w_*)$ over any $\w \in \N(\W , \varepsilon)$.
\begin{lemma} \label{lem:net} Under {\bf Assumptions~\ref{ass:2} and~\ref{ass:4}(d)}, with probability at least $1-\delta$, for any $\w \in \N(\W , \varepsilon)$, we have
\[
\left\| \nabla F(\w) - \nabla F(\w_*) -  [\nabla \Fh(\w) - \nabla \Fh(\w_*)]\right\|  \leq    \frac{LC(\varepsilon) \|\w - \w_*\| }{n}  + \sqrt{\frac{LC(\varepsilon) (F(\w)-  F(\w_*) )}{n} }.
\]
where $C(\varepsilon)$ is define in (\ref{eqn:c:eps}).
\end{lemma}
Then, we  extend the uniform convergence over $\wh$. From the property of $\varepsilon$-net, we know that there exists an point $\wt \in \N(\W , \varepsilon)$ such that $\|\wh - \wt\| \leq \varepsilon$. From the smoothness of $F(\cdot)$ and $\Fh(\cdot)$, we have
\begin{equation} \label{eqn:smooth:net}
\begin{split}
& \left\| \nabla F(\wh) - \nabla F(\w_*) - [\nabla \Fh(\wh) - \nabla \Fh(\w_*)] \right\| \\
\leq & \left\| \nabla F(\wt) - \nabla F(\w_*) - [\nabla \Fh(\wt) - \nabla \Fh(\w_*)] \right\| + 2L \varepsilon.
\end{split}
\end{equation}
Combining with Lemma~\ref{lem:net}, with probability at least $1-\delta$, we have
\begin{equation} \label{eqn:add:1}
\begin{split}
& \left\| \nabla F(\wh) - \nabla F(\w_*) - [\nabla \Fh(\wh) - \nabla \Fh(\w_*)] \right\| \\
\leq &    \frac{LC(\varepsilon) \|\wt - \w_*\| }{n}+ \sqrt{\frac{LC(\varepsilon)(F(\wt)-  F(\w_*) )}{n} } + 2L \varepsilon \\
\leq & \frac{LC(\varepsilon) \|\wh - \w_*\| }{n} + \frac{LC(\varepsilon) \varepsilon}{n}  + 2L \varepsilon \\
&+ \sqrt{\frac{LC(\varepsilon)(F(\wh)-  F(\w_*) )}{n} }  + \sqrt{\frac{LC(\varepsilon)(|F(\wh)-  F(\wt)| )}{n} } \\
\overset{\text{(\ref{eqn:F:Lipschitz})}}{\leq} & \frac{LC(\varepsilon) \|\wh - \w_*\| }{n} + \sqrt{\frac{LC(\varepsilon)(F(\wh)-  F(\w_*) )}{n} }  + \frac{LC(\varepsilon) \epsilon}{n}+ \sqrt{\frac{LC(\varepsilon) G \varepsilon}{n} } + 2L \varepsilon.
\end{split}
\end{equation}

Next, we proceed to bound $A_2$ in (\ref{eqn:main:inequality}), and develop the following lemma.
\begin{lemma} \label{eqn:vec:con} Under {\bf Assumption~\ref{ass:2}}, with probability at least $1-\delta$, we have
\begin{equation} \label{eqn:add:2}
\left\|\nabla F(\w_*) - \nabla \Fh(\w_*)\right\| \leq \frac{2M\log(2/\delta)}{n} + \sqrt{\frac{8 L F_* \log(2/\delta)}{n}}.
\end{equation}
\end{lemma}
Substituting (\ref{eqn:add:1}) and  (\ref{eqn:add:2}) into (\ref{eqn:main:inequality}), with probability at least $1-2\delta$, we have
\begin{equation} \label{eqn:add:3}
\begin{split}
& F(\wh) - F(\w_*)  \\
&\leq  \frac{LC(\varepsilon)\|\wh - \w_*\|^2 }{n} + \left\| \wh - \w_* \right\| \sqrt{\frac{LC(\varepsilon)(F(\wh)-  F(\w_*) )}{n}}   \\
& +  \frac{2M\log(2/\delta)\left\| \wh - \w_* \right\| }{n} + \left\| \wh - \w_* \right\| \sqrt{\frac{8 L F_* \log(2/\delta)}{n}} \\
&  + 2L \varepsilon \left\| \wh - \w_* \right\| +\left\| \wh - \w_* \right\| \sqrt{\frac{LC(\varepsilon) G \varepsilon}{n}}+ \frac{LC(\varepsilon) \varepsilon \|\wh - \w_*\|}{n} \\
&\overset{\text{(\ref{eqn:inequality:1}), (\ref{eqn:inequality:2})}}{\leq}    \frac{2LC(\varepsilon)\|\wh - \w_*\|^2 }{n} +    \frac{2M\log(2/\delta)\left\| \wh - \w_* \right\|}{n}  + \left\| \wh - \w_* \right\| \sqrt{\frac{8 L F_* \log(2/\delta)}{n}} \\
& +\frac{F(\wh)-  F(\w_*) }{2}+ 2L \varepsilon \left\| \wh - \w_* \right\| +\frac{G \varepsilon}{2}+ \frac{LC(\varepsilon) \varepsilon \|\wh - \w_*\|}{n}
\end{split}
\end{equation}
where the last step is due to
\begin{eqnarray}
\left\| \wh - \w_* \right\| \sqrt{\frac{LC(\varepsilon)(F(\wh)-  F(\w_*) )}{n}}  \leq \frac{LC(\varepsilon) \left\| \wh - \w_* \right\|^2}{2 n} + \frac{F(\wh)-  F(\w_*) }{2}, \label{eqn:inequality:1} \\
\left\| \wh - \w_* \right\| \sqrt{\frac{LC(\varepsilon) G \varepsilon}{n}} \leq \frac{LC(\varepsilon) \left\| \wh - \w_* \right\|^2}{ 2n} +  \frac{G \varepsilon}{2}. \label{eqn:inequality:2}
\end{eqnarray}
From (\ref{eqn:add:3}),  we get
\[
\begin{split}
& \frac{1}{2}\left( F(\wh) - F(\w_*) \right) \\
\leq &   \frac{2LC(\varepsilon)\|\wh - \w_*\|^2 }{n} +    \frac{2M\log(2/\delta)\left\| \wh - \w_* \right\|}{n}  + \left\| \wh - \w_* \right\| \sqrt{\frac{8 L F_* \log(2/\delta)}{n}} \\
&+ 2L \varepsilon \left\| \wh - \w_* \right\| +\frac{ G \varepsilon}{2}+ \frac{LC(\varepsilon) \varepsilon \|\wh - \w_*\|}{n} \\
\overset{\text{(\ref{eqn:domain:R})}}{\leq} &   \frac{8 R^2 LC(\varepsilon) }{n} +    \frac{4RM\log(2/\delta)}{n}  + 4R \sqrt{\frac{2 L F_* \log(2/\delta)}{n}}  + \left(4 R L   + \frac{G}{2}+ \frac{2 R LC(\varepsilon) }{n} \right)\varepsilon,
\end{split}
\]
which implies (\ref{eqn:smooth:r1}).

\subsection{Proof of Lemma~\ref{lem:net}}
We  introduce Lemma 2 of \cite{Smale:learning}.
\begin{lemma} \label{lem:con} Let $\H$ be a Hilbert space and let $\xi$ be a random variable with values in $\H$. Assume $\|\xi\|\leq M < \infty$ almost surely. Denote $\sigma^2(\xi)=\E\left[\|\xi\|^2\right]$. Let  $\{\xi_i\}_{i=1}^m$ be $m$ ($m < \infty$) independent drawers of $\xi$. For any $0 < \delta < 1$, with confidence $1-\delta$,
\[
\left\| \frac{1}{m} \sum_{i=1}^m \left[\xi_i -\E[\xi_i]\right] \right\| \leq \frac{2 M \log(2/\delta)}{m} + \sqrt{\frac{2 \sigma^2(\xi) \log(2/\delta)}{m}}.
\]
\end{lemma}

We first consider a fixed $\w \in \N(\W , \varepsilon)$. Since $f_i(\cdot)$ is $L$-smooth, we have
\begin{equation} \label{eqn:smooth:lemma}
\left\| \nabla f_i(\w) - \nabla f_i (\w_*) \right\| \overset{\text{(\ref{eqn:f:smooth})}}{\leq}
 L \|\w - \w_*\|.
\end{equation}
Because $f_i(\cdot)$ is both convex and $L$-smooth, by (2.1.7) of \cite{nesterov2004introductory}, we have
\[
\left\| \nabla f_i(\w) - \nabla f_i (\w_*) \right\|^2  \leq L \left(f_i (\w)-  f_i(\w_*)  - \langle \nabla f_i(\w_*), \w-\w_* \rangle \right).
\]
Taking expectation over both sides, we have
\[
 \E \left[ \left\| \nabla f_i(\w) - \nabla f_i (\w_*) \right\|^2\right]
\leq  L \left(F(\w)-  F(\w_*)  - \langle \nabla F(\w_*), \w-\w_* \rangle \right) \leq L \left(F(\w)-  F(\w_*)  \right)
\]
where the last inequality follows from the optimality condition of $\w_*$, i.e.,
\[
\langle \nabla F(\w_*) , \w - \w_* \rangle \geq 0, \ \forall \w \in \W .
\]
Following Lemma~\ref{lem:con},  with probability at least $1-\delta$, we have
\[
\begin{split}
& \left\| \nabla F(\w) - \nabla F(\w_*) - [\nabla \Fh(\w) - \nabla \Fh(\w_*)] \right\| \\
=& \left\| \nabla F(\w) - \nabla F(\w_*) - \frac{1}{n} \sum_{i=1}^n [ \nabla f_i(\w) - \nabla f_i (\w_*)] \right\| \\
 \leq & \frac{2 L \|\w - \w_*\| \log(2/\delta)}{n} + \sqrt{\frac{2 L(F(\w)-  F(\w_*) ) \log(2/\delta)}{n}}.
\end{split}
\]
We obtain Lemma~\ref{lem:net} by taking the union bound over all $\w \in \N(\W , \varepsilon)$. To this end, we need an upper bound of the covering number $|\N(\W , \varepsilon)|$.

Let $\B$ be an unit ball of $d$ dimension, and  $\N(\B, \varepsilon)$ be its $\varepsilon$-net with minimal cardinality. According to a standard volume comparison argument~\citep{Convex:body:89}, we have
\[
\log|\N(\B, \varepsilon)| \leq d\log \frac{3}{\varepsilon}.
\]
Let $\B(R)$ be a ball centered at origin with radius $R$. Since we assume $\W \subseteq \B(R)$,  it follows that 
\[
\log|\N(\W , \varepsilon)| \leq \log\left|\N\left(\B(R), \frac{\varepsilon}{2} \right)\right| \leq d\log\frac{6 R}{\varepsilon}
\]
where  the first inequality is because the covering numbers are (almost) increasing by inclusion \cite[(3.2)]{OneBit:Plan:LP}.
\subsection{Proof of Lemma~\ref{eqn:vec:con}}
To apply Lemma~\ref{lem:con}, we need an upper bound of $\E\left[\|\nabla f_i (\w_*)\|^2\right]$. Since $f_i(\cdot)$ is $L$-smooth and nonnegative, from Lemma 4.1 of \cite{Smooth:Risk}, we have
\[
\|\nabla f_i(\w_*)\|^2 \leq 4 L  f_i (\w_*)
\]
and thus
\[
\E\left[\|\nabla f_i (\w_*)\|^2\right] \leq 4 L  \E\left[ f_i (\w_*) \right]=  4 L F_*.
\]
From the definition in (\ref{eqn:defin}), we have $\|\nabla f_i (\w_*)\|\leq M $. Then, according to Lemma~\ref{lem:con}, with probability at least $1-\delta$, we have
\[
 \left\|\nabla F(\w_*) - \nabla \Fh(\w_*)\right\| = \left\|\nabla F(\w_*) - \frac{1}{n}\sum_{i=1}^n \nabla f_i(\w_*) \right\|\leq  \frac{2M\log(2/\delta)}{n} + \sqrt{\frac{8 L F_* \log(2/\delta)}{n}}.
\]

\subsection{Proof of Theorem~\ref{thm:smooth:convex}}
The proof follows the same logic as that of Theorem~\ref{thm:smooth}. Under {\bf Assumption~\ref{ass:4}(b)}, (\ref{eqn:main:inequality}) becomes
\begin{equation} \label{eqn:main:inequality:2}
\begin{split}
& F(\wh) - F(\w_*) + \frac{\lambda}{2} \|\wh-\w_*\|^2\\
\leq &\left( \underbrace{\left\| \nabla F(\wh) - \nabla F(\w_*) - [\nabla \Fh(\wh) - \nabla \Fh(\w_*)] \right\| }_{:=A_1}+ \underbrace{\left\| \nabla F(\w_*) - \nabla \Fh(\w_*) \right\|}_{:=A_2} \right)\left\| \wh - \w_* \right\|.
\end{split}
\end{equation}
Substituting (\ref{eqn:add:1}) and  (\ref{eqn:add:2}) into (\ref{eqn:main:inequality:2}), with probability at least $1-2\delta$, we have
\begin{equation} \label{eqn:strongly:convex}
\begin{split}
& F(\wh) - F(\w_*) + \frac{\lambda}{2} \|\wh-\w_*\|^2  \\
&\leq  \frac{LC(\varepsilon)\|\wh - \w_*\|^2 }{n} + \left\| \wh - \w_* \right\| \sqrt{\frac{LC(\varepsilon)(F(\wh)-  F(\w_*) )}{n}}  \\
& +  \frac{2M\log(2/\delta)\left\| \wh - \w_* \right\| }{n} + \left\| \wh - \w_* \right\| \sqrt{\frac{8 L F_* \log(2/\delta)}{n}} \\
& + 2L \varepsilon \left\| \wh - \w_* \right\| +\left\| \wh - \w_* \right\| \sqrt{\frac{LC(\varepsilon) G \varepsilon}{n}}+ \frac{LC(\varepsilon) \varepsilon \|\wh - \w_*\|}{n} \\
\end{split}
\end{equation}

To prove (\ref{eqn:main:1}), we substitute (\ref{eqn:inequality:1}), (\ref{eqn:inequality:2}), and
\[
\left\| \wh - \w_* \right\| \sqrt{\frac{8 L F_* \log(2/\delta)}{n}} \leq  \frac{4 L F_* \log(2/\delta)}{\lambda n} + \frac{\lambda}{2} \left\| \wh - \w_* \right\|^2
\]
into (\ref{eqn:strongly:convex}), and then obtain
\[
\begin{split}
& \frac{1}{2} \left( F(\wh) - F(\w_*) \right) \\
&\leq    \frac{2LC(\varepsilon)\|\wh - \w_*\|^2 }{n} +    \frac{2M\log(2/\delta)\left\| \wh - \w_* \right\|}{n}  + \frac{4 L F_* \log(2/\delta)}{\lambda n}\\
& + 2L \varepsilon \left\| \wh - \w_* \right\|+\frac{G \varepsilon}{2} + \frac{LC(\varepsilon) \varepsilon \|\wh - \w_*\|}{n} \\
&\overset{\text{(\ref{eqn:domain:R})}}{\leq}    \frac{8 R^2 LC(\varepsilon) }{n} +    \frac{4 RM\log(2/\delta)}{n}  + \frac{4 L F_* \log(2/\delta)}{\lambda n} + \left( 4 R L +  \frac{G }{2}+ \frac{2 R LC(\varepsilon) }{n} \right)\varepsilon.
\end{split}
\]
which implies (\ref{eqn:main:1}).

To prove (\ref{eqn:main:2}), we substitute
\[
\begin{split}
\left\| \wh - \w_* \right\| \sqrt{\frac{LC(\varepsilon)(F(\wh)-  F(\w_*) )}{n}} \leq \frac{2 LC(\varepsilon)(F(\wh)-  F(\w_*) )}{\lambda n} +\frac{\lambda}{8} \left\| \wh - \w_* \right\|^2,   \\
\frac{2M\log(2/\delta)\left\| \wh - \w_* \right\|}{n} \leq  \frac{16M^2\log^2(2/\delta)}{\lambda n^2} + \frac{\lambda}{16} \left\| \wh - \w_* \right\|^2,  \\
\left\| \wh - \w_* \right\| \sqrt{\frac{8 L F_* \log(2/\delta)}{n}} \leq \frac{64 L F_* \log(2/\delta)}{\lambda n} +\frac{\lambda}{32} \left\| \wh - \w_* \right\|^2, \\
2L \varepsilon \left\| \wh - \w_* \right\| \leq   \frac{64 L^2 \varepsilon^2}{\lambda} +  \frac{\lambda}{64} \left\| \wh - \w_* \right\|^2,\\
\left\| \wh - \w_* \right\| \sqrt{\frac{LC(\varepsilon) G \varepsilon}{n}} \leq  \frac{32 LC(\varepsilon) G \varepsilon}{ \lambda n}+\frac{\lambda}{128} \left\| \wh - \w_* \right\|^2, \\
\frac{LC(\varepsilon) \varepsilon \|\wh - \w_*\|}{n}  \leq \frac{32 L^2 C^2(\varepsilon) \varepsilon^2}{\lambda n^2} + \frac{\lambda}{128} \left\| \wh - \w_* \right\|^2
\end{split}
\]
into (\ref{eqn:strongly:convex}), and then obtain
\[
\begin{split}
& F(\wh) - F(\w_*) + \frac{\lambda}{4} \left\| \wh - \w_* \right\|^2 \\
\leq &   \frac{LC(\varepsilon)\|\wh - \w_*\|^2 }{ n}  + \frac{2 LC(\varepsilon)(F(\wh)-  F(\w_*) )}{\lambda n} +  \frac{16M^2\log^2(2/\delta)}{\lambda n^2}+  \frac{64 L F_* \log(2/\delta)}{\lambda n} \\
& + \frac{64 L^2 \varepsilon^2}{\lambda}+\frac{32 LC(\varepsilon) G \varepsilon}{ \lambda n}  +\frac{32 L^2 C^2(\varepsilon) \varepsilon^2}{\lambda n^2} \\
\overset{\text{(\ref{eqn:lower:n})}}{\leq} & \frac{\lambda}{4} \left\| \wh - \w_* \right\|^2 + \frac{1}{2}\left( F(\wh)-  F(\w_*) \right) +  \frac{16M^2\log^2(2/\delta)}{\lambda n^2}+  \frac{64 L F_* \log(2/\delta)}{\lambda n}\\
& + \frac{64 L^2 \varepsilon^2}{\lambda}+8 G \varepsilon  +2 \lambda \varepsilon^2\\
\end{split}
\]
which implies (\ref{eqn:main:2}).
\subsection{Proof of Theorem~\ref{thm:smooth:convex:2}}
Without {\bf Assumption~\ref{ass:4}(d)}, Lemma~\ref{lem:net} which is used in the proofs of Theorems~\ref{thm:smooth} and \ref{thm:smooth:convex} does not hold anymore. Instead, we will use the following version that only relies on the smoothness condition.
\begin{lemma} \label{lem:net:new} Under {\bf Assumption~\ref{ass:2}}, with probability at least $1-\delta$, for any $\w \in \N(\W , \varepsilon)$, we have
\[
\left\| \nabla F(\w) - \nabla F(\w_*) -  [\nabla \Fh(\w) - \nabla \Fh(\w_*)]\right\|  \leq    \frac{LC(\varepsilon) \|\w - \w_*\| }{n}  + L \|\w - \w_*\|\sqrt{\frac{C(\varepsilon) }{n} }
\]
where $C(\varepsilon)$ is define in (\ref{eqn:c:eps}).
\end{lemma}
The above lemma is a direct consequence of (\ref{eqn:smooth:lemma}), Lemma~\ref{lem:con} and the union bound.

The rest of the proof is similar to those of Theorems~\ref{thm:smooth} and \ref{thm:smooth:convex}. We first derive a counterpart of (\ref{eqn:add:1}) under Lemma~\ref{lem:net:new}. Combining (\ref{eqn:smooth:net}) with Lemma~\ref{lem:net:new}, with probability at least $1-\delta$, we have
\begin{equation} \label{eqn:add:1:new}
\begin{split}
& \left\| \nabla F(\wh) - \nabla F(\w_*) - [\nabla \Fh(\wh) - \nabla \Fh(\w_*)] \right\| \\
\leq &    \frac{LC(\varepsilon) \|\wt - \w_*\| }{n}+ L \|\wt - \w_*\|\sqrt{\frac{C(\varepsilon) }{n} } + 2L \varepsilon \\
\leq & \frac{LC(\varepsilon) \|\wh - \w_*\| }{n}   + L \|\wh - \w_*\|\sqrt{\frac{C(\varepsilon) }{n} }  + \frac{LC(\varepsilon) \varepsilon}{n}+ L \varepsilon\sqrt{\frac{C(\varepsilon) }{n} } + 2L \varepsilon .
\end{split}
\end{equation}
Substituting (\ref{eqn:add:1:new}) and  (\ref{eqn:add:2}) into (\ref{eqn:main:inequality:2}), with probability at least $1-2\delta$, we have
\begin{equation} \label{eqn:strongly:convex:new}
\begin{split}
& F(\wh) - F(\w_*) + \frac{\lambda}{2} \|\wh-\w_*\|^2  \\
\leq & \frac{LC(\varepsilon)\|\wh - \w_*\|^2 }{n} + L \|\wh - \w_*\|^2\sqrt{\frac{C(\varepsilon) }{n} }  \\
& +  \frac{2M\log(2/\delta)\left\| \wh - \w_* \right\| }{n} + \left\| \wh - \w_* \right\| \sqrt{\frac{8 L F_* \log(2/\delta)}{n}} \\
& + 2L \varepsilon \left\| \wh - \w_* \right\|+ L \varepsilon \left\| \wh - \w_* \right\|\sqrt{\frac{C(\varepsilon) }{n} }+\frac{LC(\varepsilon) \varepsilon \|\wh - \w_*\|}{n}  .
\end{split}
\end{equation}
To get (\ref{eqn:main:1:2}), we substitute
\[
\begin{split}
 L \|\wh - \w_*\|^2\sqrt{\frac{C(\varepsilon) }{n} } \leq  \frac{L^2 C(\varepsilon) \left\| \wh - \w_* \right\|^2 }{\lambda n} + \frac{\lambda}{4} \left\| \wh - \w_* \right\|^2,\\
\left\| \wh - \w_* \right\| \sqrt{\frac{8 L F_* \log(2/\delta)}{n}} \leq  \frac{8 L F_* \log(2/\delta)}{\lambda n} + \frac{\lambda}{4} \left\| \wh - \w_* \right\|^2
\end{split}
\]
into (\ref{eqn:strongly:convex:new}), and then obtain
\[
\begin{split}
& F(\wh) - F(\w_*)  \\
\leq & \frac{LC(\varepsilon)\|\wh - \w_*\|^2 }{n} + \frac{L^2 C(\varepsilon) \left\| \wh - \w_* \right\|^2 }{\lambda n}  +  \frac{2M\log(2/\delta)\left\| \wh - \w_* \right\| }{n} + \frac{8 L F_* \log(2/\delta)}{\lambda n}  \\
&  + 2L \varepsilon \left\| \wh - \w_* \right\|+ L \varepsilon \left\| \wh - \w_* \right\| \sqrt{\frac{C(\varepsilon) }{n} } + \frac{LC(\varepsilon) \varepsilon \|\wh - \w_*\|}{n} \\
\overset{\text{(\ref{eqn:domain:R})}}{\leq} &  \frac{4R^2LC(\varepsilon)}{n} + \frac{4 R^2 L^2 C(\varepsilon) }{\lambda n}  +  \frac{4 RM\log(2/\delta) }{n} + \frac{8 L F_* \log(2/\delta)}{\lambda n} \\
& +\left( 4  R L  +2 R L  \sqrt{\frac{C(\varepsilon) }{n} } + \frac{2 R L C(\varepsilon) }{n} \right) \varepsilon
\end{split}
\]
which proves (\ref{eqn:main:1:2}).

To get (\ref{eqn:main:2:2}), we  substitute
\[
\begin{split}
\frac{2M\log(2/\delta)\left\| \wh - \w_* \right\|}{n} \leq  \frac{8M^2\log^2(2/\delta)}{\lambda n^2} + \frac{\lambda}{8} \left\| \wh - \w_* \right\|^2,  \\
\left\| \wh - \w_* \right\| \sqrt{\frac{8 L F_* \log(2/\delta)}{n}} \leq \frac{32 L F_* \log(2/\delta)}{\lambda n} +\frac{\lambda}{16} \left\| \wh - \w_* \right\|^2, \\
2L   \varepsilon \left\| \wh - \w_* \right\|\leq   \frac{32 L^2 \varepsilon^2}{\lambda} +  \frac{\lambda}{32} \left\| \wh - \w_* \right\|^2,\\
 L \varepsilon \left\| \wh - \w_* \right\| \sqrt{\frac{C(\varepsilon) }{n} } \leq  \frac{16 L^2C(\varepsilon) \varepsilon^2}{ \lambda n}+\frac{\lambda}{64} \left\| \wh - \w_* \right\|^2, \\
\frac{LC(\varepsilon) \varepsilon \|\wh - \w_*\|}{n}  \leq \frac{16 L^2 C^2(\varepsilon) \varepsilon^2}{\lambda n^2} + \frac{\lambda}{64} \left\| \wh - \w_* \right\|^2
\end{split}
\]
into (\ref{eqn:strongly:convex:new}), and then obtain
\[
\begin{split}
& F(\wh) - F(\w_*) + \frac{\lambda}{4} \|\wh-\w_*\|^2  \\
\leq & \frac{LC(\varepsilon)\|\wh - \w_*\|^2 }{n} + L \|\wh - \w_*\|^2\sqrt{\frac{C(\varepsilon) }{n} }  +    \frac{8M^2\log^2(2/\delta)}{\lambda n^2} + \frac{32 L F_* \log(2/\delta)}{\lambda n} \\
& + \left(\frac{32 L^2}{\lambda} +  \frac{16 L^2C(\varepsilon)}{ \lambda n} + \frac{16 L^2 C^2(\varepsilon) }{\lambda n^2}\right)\varepsilon^2\\
\overset{\text{(\ref{eqn:lower:n:2})}}{\leq} & \frac{\lambda^2\|\wh - \w_*\|^2 }{25 L } + \frac{\lambda}{5} \|\wh - \w_*\|^2  + \frac{8M^2\log^2(2/\delta)}{\lambda n^2} + \frac{32 L F_* \log(2/\delta)}{\lambda n} \\
& + \left(\frac{32 L^2}{\lambda} +  \frac{16 \lambda }{25} + \frac{16 \lambda^3  }{625 L^2}\right)\varepsilon^2\\
\overset{\lambda/L \leq 1}{\leq} & \frac{6\lambda }{25}\|\wh - \w_*\|^2  + \frac{8M^2\log^2(2/\delta)}{\lambda n^2} + \frac{32 L F_* \log(2/\delta)}{\lambda n}  + \left(\frac{32 L^2}{\lambda}  + \frac{416 \lambda  }{625}\right)\varepsilon^2.
\end{split}
\]
By subtracting $\lambda\|\wh - \w_*\|^2/4$ from both sides we complete the proof of (\ref{eqn:main:2:2}).

\subsection{Proof of Theorem \ref{thm:sup:learn}}
We consider two cases. In the first case, we assume that
\[
\|\wh - \w_*\| \leq \frac{1}{n^2}.
\]
Since $H(\cdot)$ is $L$-smooth and $r(\cdot)$ is $P$-Lipschitz continuous, we have
\begin{equation} \label{eqn:main:first}
\begin{split}
&  F(\wh) - F(\w_*) = H(\wh) + r(\wh) - H(\w_*)-r(\w_*)\\
\leq & \langle \wh - \w_*, \nabla H(\w_*) \rangle + \frac{L}{2}\|\wh - \w_*\|^2 + P \|\wh - \w_*\| \\
\leq &¡¡ \| \wh - \w_*\| \|\nabla H(\w_*)\| + \frac{L}{2}\|\wh - \w_*\|^2 + P \|\wh - \w_*\| \leq \frac{M +P }{n^2}+ \frac{L}{2n^4}
\end{split}
\end{equation}
where the last step utilizes Jensen's inequality
\[
\|\nabla H(\w_*)\| =  \left\| \E_{(\x,y) \sim \D} \left[ \nabla \ell(\langle \w_*, \x\rangle, y) \right]\right \| \leq \E_{(\x,y) \sim \D}\left[ \left\| \nabla \ell(\langle \w_*, \x\rangle, y)\right \| \right] \overset{\text{(\ref{eqn:sup:M})}}{\leq} M.
\]
Next, we study the case
\[
\frac{1}{n^2}  <  \|\wh - \w_*\| \overset{\text{(\ref{eqn:domain:R:New})}}{\leq} 2R.
\]
From (\ref{eqn:key}),  we have
 \begin{equation} \label{eqn:main:inequality:sup}
\begin{split}
& F(\wh) - F(\w_*) + \frac{\lambda}{2} \|\wh-\w_*\|^2\\
\leq  &  \langle \nabla F(\wh) - \nabla F(\w_*) - [\nabla \Fh(\wh) - \nabla \Fh(\w_*)], \wh - \w_* \rangle + \langle \nabla F(\w_*) - \nabla \Fh(\w_*), \wh - \w_*\rangle \\
=& \langle \nabla H(\wh) - \nabla H(\w_*) - [\nabla \Hh(\wh) - \nabla \Hh(\w_*)], \wh - \w_* \rangle + \langle \nabla H(\w_*) - \nabla \Hh(\w_*), \wh - \w_*\rangle \\
\leq & \underbrace{\sup\limits_{\w:\|\w - \w_*\| \leq \|\wh - \w_*\|} \left\langle \nabla H(\w) - \nabla H(\w_*) - [\nabla \Hh(\w) - \nabla \Hh(\w_*)], \w - \w_* \right\rangle}_{:=B_1} \\
&+\underbrace{\left\| \nabla H(\w_*) - \nabla \Hh(\w_*) \right\|}_{:=B_2} \left\| \wh - \w_* \right\|.
\end{split}
\end{equation}
We first bound $B_1$. To utilize the fact the random variable $\|\wh - \w_*\|$ lies in the range $(1/n^2, 2R]$, we develop the following lemma.
\begin{lemma} \label{lem:union} Under {\bf Assumptions~\ref{ass:5}} and {\bf~\ref{ass:6}}, with probability at least $1-\delta$, for all
\[
 \frac{1}{n^2}< \gamma \leq 2R
\]
the following bound holds:
\[
\sup\limits_{\w:\|\w - \w_*\| \leq \gamma} \left\langle \nabla H(\w) - \nabla H(\w_*) - [\nabla \Hh(\w) - \nabla \Hh(\w_*)], \w - \w_* \right\rangle \leq \frac{4 L \gamma^2  }{\sqrt{n}} \left( 8 +  \sqrt{2 \log \frac{s}{\delta}} \right)
\]
where $s= \lceil 2 \log_2(n) + \log_2(2R)\rceil$.
\end{lemma}
Based on the above lemma, we have with probability at least $1-\delta$,
\begin{equation}\label{eqn:B1}
B_1 \leq \frac{4 L \|\wh - \w_*\|^2 }{\sqrt{n}} \left( 8 +  \sqrt{2 \log \frac{s}{\delta}} \right) = \frac{L C \|\wh - \w_*\|^2 }{\sqrt{n}}
\end{equation}
where $C$ is defined in (\ref{eqn:defin:C}).

We then proceed to handle $B_2$, which can be upper bounded in the same way as $A_2$. In particular, we have the following lemma.
\begin{lemma}  Under {\bf Assumptions~\ref{ass:5}} and {\bf\ref{ass:6}}, with probability at least $1-\delta$, we have
\begin{equation} \label{eqn:B2}
\left\|\nabla H(\w_*) - \nabla \Hh(\w_*)\right\| \leq \frac{2M\log(2/\delta)}{n} + \sqrt{\frac{8 L H_* \log(2/\delta)}{n}}.
\end{equation}
\end{lemma}

Substituting (\ref{eqn:B1}) and  (\ref{eqn:B2}) into (\ref{eqn:main:inequality:sup}), with probability at least $1-2\delta$, we have
\begin{equation} \label{eqn:sup:key}
\begin{split}
& F(\wh) - F(\w_*) + \frac{\lambda}{2} \|\wh-\w_*\|^2\\
\leq  & \frac{LC\|\wh - \w_*\|^2 }{\sqrt{n}}  + \frac{2M\log(2/\delta)\| \wh - \w_* \|}{n} +\| \wh - \w_* \| \sqrt{\frac{8 L H_* \log(2/\delta)}{n}} .
\end{split}
\end{equation}
We substitute
\[
\begin{split}
\frac{LC\|\wh - \w_*\|^2 }{\sqrt{n}} \leq  \frac{L^2C^2 \|\wh - \w_*\|^2}{\lambda n} + \frac{\lambda}{4}\|\wh - \w_*\|^2,\\
\| \wh - \w_* \| \sqrt{\frac{8 L H_* \log(2/\delta)}{n}}\leq  \frac{8 L H_* \log(2/\delta)}{\lambda n} + \frac{\lambda}{4}\|\wh - \w_*\|^2
\end{split}
\]
into (\ref{eqn:sup:key}), and then have
\[
\begin{split}
F(\wh) - F(\w_*)\leq& \frac{L^2C^2 \|\wh - \w_*\|^2}{\lambda n}+ \frac{2M\log(2/\delta)\| \wh - \w_* \|}{n} +\frac{8 L H_* \log(2/\delta)}{\lambda n}\\
\overset{\text{(\ref{eqn:domain:R})}}{\leq} & \frac{4 R^2 L^2 C^2 }{\lambda n}+ \frac{4 RM\log(2/\delta)}{n}  +\frac{8 L H_* \log(2/\delta)}{\lambda n}.
\end{split}
\]
Combining the above inequality with (\ref{eqn:main:first}), we obtain  (\ref{eqn:sup:1}).

To prove (\ref{eqn:sup:2}), we substitute
\[
\begin{split}
\frac{2M\log(2/\delta)\| \wh - \w_* \|}{n} \leq \frac{8M^2\log^2(2/\delta)}{\lambda n^2} +\frac{\lambda}{8}\|\wh - \w_*\|^2,\\
\| \wh - \w_* \| \sqrt{\frac{8 L H_* \log(2/\delta)}{n}}\leq  \frac{16 L H_* \log(2/\delta)}{\lambda n} + \frac{\lambda}{8}\|\wh - \w_*\|^2
\end{split}
\]
into (\ref{eqn:sup:key}), and then have
\[
\begin{split}
& F(\wh) - F(\w_*) +\frac{\lambda}{4}\|\wh - \w_*\|^2 \\
\leq& \frac{L C\|\wh - \w_*\|^2 }{\sqrt{n}}+ \frac{8M^2\log^2(2/\delta)}{\lambda n^2}+\frac{16 L H_* \log(2/\delta)}{\lambda n}\\
\overset{\text{(\ref{eqn:sup:lower:n})}}{\leq} & \frac{\lambda}{4}\|\wh - \w_*\|^2+ \frac{8M^2\log^2(2/\delta)}{\lambda n^2}+\frac{16 L H_* \log(2/\delta)}{\lambda n}.
\end{split}
\]
Combining the above inequality with (\ref{eqn:main:first}), we obtain  (\ref{eqn:sup:2}).

\section{Conclusions and Future work}
In this paper, we study the excess risk of ERM for SCO. Our theoretical results show that it is possible to achieve $O(1/n)$-type of risk bounds under (i) the smoothness and small minimal risk  conditions (i.e., Theorem~\ref{thm:smooth}) or (ii) the smoothness and strong convexity conditions (i.e., the first part of Theorems~\ref{thm:smooth:convex}, \ref{thm:smooth:convex:2}, and \ref{thm:sup:learn}). A more exciting result is that when $n$ is large enough, ERM has  $O(1/n^2)$-type of risk bounds under the smoothness, strong convexity, and small minimal risk conditions (i.e., the second part of Theorems~\ref{thm:smooth:convex}, \ref{thm:smooth:convex:2}, and \ref{thm:sup:learn}).

In the context of SCO, there remain many open problems about ERM.
\begin{compactenum}
\item Our current results are restricted to the Hilbert or Euclidean space, because the smoothness and strong convexity are defined in terms of the $\ell_2$-norm. We will extend our analysis to other geometries in the future.
\item As  mentioned in {\bf Remark 3}, under the strong convexity condition, a dimensionality-independent   risk bound, e.g., $\O(\kappa/ n)$ or $\O(1/ \lambda n)$, that holds with high probability is still missing.
\item As discussed in {\bf Remark 9}, it is unclear whether the convexity of the loss can be exploited to improve the lower bound of $n$ in the second part of Theorem~\ref{thm:sup:learn}. Ideally, we expect that $n=\Omega(\kappa)$ is sufficient to deliver an $O(1/[\lambda n^2] + \kappa H_*/n)$ risk bound.
\item The  $O(1/n^2)$-type of risk bounds require both the smoothness and strong convexity conditions. One may investigate  whether  strong convexity  can be relaxed to other weaker conditions, such as exponential concavity.
\end{compactenum}
Finally, as far as we know, there are no $O(1/n^2)$-type of risk bounds for stochastic approximation (SA). We will try to establish such bounds for SA.


\bibliography{ref}
\appendix

\section{Proof of Theorem \ref{thm:smooth:weak}}
This result is actually a  byproduct of Theorem~\ref{thm:smooth:convex:2}. Since strong convexity is absent, we set $\lambda=0$ in (\ref{eqn:strongly:convex:new}) and obtain
\[
\begin{split}
& F(\wh) - F(\w_*) \\
\leq & \frac{LC(\varepsilon)\|\wh - \w_*\|^2 }{n} + L \|\wh - \w_*\|^2\sqrt{\frac{C(\varepsilon) }{n} }  \\
& +  \frac{2M\log(2/\delta)\left\| \wh - \w_* \right\| }{n} + \left\| \wh - \w_* \right\| \sqrt{\frac{8 L F_* \log(2/\delta)}{n}} \\
& + 2L \varepsilon \left\| \wh - \w_* \right\|+ L \varepsilon \left\| \wh - \w_* \right\|\sqrt{\frac{C(\varepsilon) }{n} }+\frac{LC(\varepsilon) \varepsilon \|\wh - \w_*\|}{n}\\
\overset{\text{(\ref{eqn:domain:R})}}{\leq} & \frac{4R^2LC(\varepsilon) }{n}+  \frac{4RM\log(2/\delta) }{n} + 4R^2 L \sqrt{\frac{C(\varepsilon) }{n} }   + 2R \sqrt{\frac{8 L F_* \log(2/\delta)}{n}} \\
& +\left( 4  R L  +2 R L  \sqrt{\frac{C(\varepsilon) }{n} } + \frac{2 R L C(\varepsilon) }{n} \right) \varepsilon.
\end{split}
\]

\section{Proof of Lemma~\ref{lem:union}}
First, we partition the range $(1/n^2, 2R]$ into $s= \lceil 2 \log_2(n) + \log_2(2R) \rceil$ consecutive segments $\Delta_1, \Delta_2, \ldots, \Delta_s$ such that
\[
\Delta_k=\left(\underbrace{\frac{2^{k-1}}{n^2} }_{:=\gamma_k^-},  \underbrace{\frac{2^{k}}{n^2} }_{:=\gamma_k^+} \right], \ k=1, \ldots, s.
\]
Then, we consider the case $\gamma \in \Delta_k$ for a fixed value of $k$. We have
\begin{equation} \label{eqn:lem:union:1}
\begin{split}
& \sup\limits_{\w:\|\w - \w_*\| \leq \gamma} \left\langle \nabla H(\w) - \nabla H(\w_*) - [\nabla \Hh(\w) - \nabla \Hh(\w_*)], \w - \w_* \right\rangle \\
\leq & \sup\limits_{\w:\|\w - \w_*\| \leq \gamma_k^+} \left\langle \nabla H(\w) - \nabla H(\w_*) - [\nabla \Hh(\w) - \nabla \Hh(\w_*)], \w - \w_* \right\rangle.
\end{split}
\end{equation}
Based on the McDiarmid's inequality \citep{McDiarmid} and the Rademacher complexity \citep{Rademacher_Gaussian}, we have the following lemma to upper bound the last term.
\begin{lemma} \label{lem:McD} Under {\bf Assumptions~\ref{ass:5}} and {\bf\ref{ass:6}}, with probability at least $1-\delta$, we have
\begin{equation} \label{eqn:lem:union:2}
\begin{split}
&\sup\limits_{\w:\|\w - \w_*\| \leq \gamma_k^+} \left\langle \nabla H(\w) - \nabla H(\w_*) - [\nabla \Hh(\w) - \nabla \Hh(\w_*)], \w - \w_* \right\rangle \\
\leq & \frac{L \left(\gamma_k^+\right)^2 }{\sqrt{n}} \left( 8 +  \sqrt{2 \log \frac{1}{\delta}} \right).
\end{split}
\end{equation}
\end{lemma}
Since $\gamma \in \Delta_k$, we have
\begin{equation} \label{eqn:lem:union:3}
\gamma_k^+ =  2 \gamma_k^- \leq 2 \gamma.
\end{equation}
 Thus, with  probability at least $1-\delta$, we have
\[
\begin{split}
& \sup\limits_{\w:\|\w - \w_*\| \leq \gamma} \left\langle \nabla H(\w) - \nabla H(\w_*) - [\nabla \Hh(\w) - \nabla \Hh(\w_*)], \w - \w_* \right\rangle \\
\overset{\text{(\ref{eqn:lem:union:1}),(\ref{eqn:lem:union:2}),(\ref{eqn:lem:union:3})}}{\leq}  & \frac{4 L \gamma^2}{\sqrt{n}} \left(8 +  \sqrt{2 \log \frac{1}{\delta}} \right).
\end{split}
\]
We complete the proof by taking the union bound over $s$ segments.
\section{Proof of Lemma~\ref{lem:McD}} \label{sec:lem:McD}
To simplify the notation, we define
\[
\begin{split}
h_i(\w) =  &\ell(\langle \w, \x_i\rangle, y_i), \ i=1,\ldots,n,\\
l(h_1,\ldots,h_n)=&\sup\limits_{\w:\|\w - \w_*\| \leq \gamma_k^+} \left\langle \nabla H(\w) - \nabla H(\w_*) - \frac{1}{n} \sum_{i=1}^n [\nabla h_i(\w) - \nabla h_i(\w_*)], \w - \w_* \right\rangle.
\end{split}
\]
To upper bound $l(h_1,\ldots,h_n)$, we utilize the McDiarmid's inequality \citep{McDiarmid}.
\begin{thm} Let $X_1,\ldots,X_n$ be independent random variables taking values in a set $A$, and assume that $f:A^n \mapsto \R$ satisfies
\[
\sup_{x_1,\ldots,x_n,x_i'\in A}\left| H(x_1,\ldots,x_n) - H(x_1,\ldots,x_{i-1},x_i',x_{i+1},\ldots,x_n) \right| \leq c_i
\]
for every $1 \leq i \leq n$. Then, for every $t >0$,
\[
P\left\{ H(X_1,\ldots,X_n) - \E \left[ H(X_1,\ldots,X_n) \right] \geq t\right\} \leq \exp\left(-\frac{2 t^2}{\sum_{i=1}^n c_i^2} \right).
\]
\end{thm}

As pointed out in Remark 7, {\bf Assumptions~\ref{ass:5}} and {\bf\ref{ass:6}} imply the random function $h_i(\cdot)$ is $L$-smooth, and thus
\[
\left| \langle \nabla h_i(\w) - \nabla h_i(\w_*), \w - \w_* \rangle \right| \leq L \|\w - \w_*\|^2 \leq L \left(\gamma_k^+\right)^2.
\]
As a result, when a random function $h_i$ changes, the random variable $l(h_1,\ldots,h_n)$ can change by no more than $2 L \left(\gamma_k^+\right)^2/n$.  McDiarmid's inequality implies that with probability at least $1-\delta$
\begin{equation} \label{eqn:McD:1}
l(h_1,\ldots,h_n) \leq \E \left[l(h_1,\ldots,h_n)\right] + L \left(\gamma_k^+\right)^2  \sqrt{\frac{2}{n} \log \frac{1}{\delta}}.
\end{equation}

Let $(h_1', \ldots, h_n')$ be an independent copy of $(h_1,\ldots,h_n)$, and  $\epsilon_1, \ldots, \epsilon_n$ be $n$ i.i.d.~Rademacher variables with equal probability of being $\pm 1$. Using techniques of Rademacher complexities \citep{Rademacher_Gaussian}, we bound $\E \left[l(h_1,\ldots,h_n)\right]$ as follows:
\[
\begin{split}
& \E_{h_1,\ldots,h_n} \left[ \sup\limits_{\w:\|\w - \w_*\| \leq \gamma_k^+} \left\langle \nabla H(\w) - \nabla H(\w_*) -  \frac{1}{n} \sum_{i=1}^n [\nabla h_i(\w) - \nabla h_i(\w_*)], \w - \w_* \right\rangle \right] \\
=& \frac{1}{n} \E_{h_1,\ldots,h_n} \left[ \sup\limits_{\w:\|\w - \w_*\| \leq \gamma_k^+} \right. \\
& \quad \left. \E_{h_1',\ldots,h_n'} \left[\sum_{i=1}^n  \left\langle  \nabla h_i'(\w) - \nabla h_i'(\w_*), \w - \w_* \right\rangle\right]  - \sum_{i=1}^n\left\langle  \nabla h_i(\w) - \nabla h_i(\w_*), \w - \w_* \right\rangle \right] \\
\leq & \frac{1}{n} \E_{h_1,\ldots,h_n, h_1',\ldots,h_n'} \left[ \sup\limits_{\w:\|\w - \w_*\| \leq \gamma_k^+}   \right.\\
& \quad \left. \sum_{i=1}^n\left\langle  \nabla h_i'(\w) - \nabla h_i'(\w_*), \w - \w_* \right\rangle - \sum_{i=1}^n\left\langle  \nabla h_i(\w) - \nabla h_i(\w_*), \w - \w_* \right\rangle \right] \\
=& \frac{1}{n} \E_{h_1,\ldots,h_n, h_1',\ldots,h_n',\epsilon_1, \ldots, \epsilon_n} \left[ \sup\limits_{\w:\|\w - \w_*\| \leq \gamma_k^+}  \right.\\
& \quad \left.\sum_{i=1}^n \epsilon_i \left( \left\langle  \nabla h_i'(\w) - \nabla h_i'(\w_*), \w - \w_* \right\rangle - \left\langle  \nabla h_i(\w) - \nabla h_i(\w_*), \w - \w_* \right\rangle \right)\right] \\
\leq & \frac{2}{n}\E_{h_1,\ldots,h_n, \epsilon_1, \ldots, \epsilon_n} \left[\sup\limits_{\w:\|\w - \w_*\| \leq \gamma_k^+}   \sum_{i=1}^n \epsilon_i \left\langle  \nabla h_i(\w) - \nabla h_i(\w_*), \w - \w_* \right\rangle \right] .
\end{split}
\]
Substituting the above inequality into (\ref{eqn:McD:1}), we obtain
\begin{equation} \label{eqn:McD:2}
\begin{split}
& l(h_1,\ldots,h_n) \\
\leq& L \left(\gamma_k^+\right)^2  \sqrt{\frac{2}{n} \log \frac{1}{\delta}}+ \frac{2}{n}\E \left[\sup\limits_{\w:\|\w - \w_*\| \leq \gamma_k^+}   \sum_{i=1}^n \epsilon_i \left\langle  \nabla h_i(\w) - \nabla h_i(\w_*), \w - \w_* \right\rangle \right].
\end{split}
\end{equation}
To upper bound the last term of (\ref{eqn:McD:2}), we use the Rademacher complexity of the product of two functions \citep{Desalvo:2015:LDC}, and develop the following lemma.
\begin{lemma} \label{lem:product}
\[
\E \left[\sup\limits_{\w:\|\w - \w_*\| \leq \gamma_k^+} \sum_{i=1}^n \epsilon_i \left\langle  \nabla h_i(\w) - \nabla h_i(\w_*), \w - \w_* \right\rangle \right] \leq  4 L \left(\gamma_k^+\right)^2   \sqrt{n}.
\]
\end{lemma}
We complete the proof by substituting the above inequality into (\ref{eqn:McD:2}).
\section{Proof of Lemma~\ref{lem:product}}
Define
\[
\begin{split}
p_i(\w)= &\frac{1}{\sqrt{\beta}} \left(\ell'(\langle \w, \x_i\rangle, y_i)-\ell'(\langle \w_*, \x_i\rangle, y_i) \right)\in [-\gamma_k^+  D \sqrt{\beta}, \gamma_k^+ D \sqrt{\beta}], \\
q_i(\w)=&\sqrt{\beta} \langle \x_i, \w-\w_* \rangle \in [-\gamma_k^+ D \sqrt{\beta}, \gamma_k^+ D \sqrt{\beta}]
\end{split}
\]
such that
\[
\begin{split}
 & \left\langle  \nabla h_i(\w) - \nabla h_i(\w_*), \w - \w_* \right\rangle=\left\langle  \nabla \ell(\langle \w, \x_i\rangle, y_i)) - \nabla \ell(\langle \w_*, \x_i\rangle, y_i), \w - \w_* \right\rangle  \\
 =&\left(\ell'(\langle \w, \x_i\rangle, y_i)-\ell'(\langle \w_*, \x_i\rangle, y_i) \right)  \langle \x_i, \w-\w_* \rangle= p_i(\w) q_i(\w).
\end{split}
\]
From the equality $ab=\frac{1}{4}\left((a+b)^2-(a-b)^2\right)$, we have
\begin{equation} \label{eqn:McD:3}
\begin{split}
&\E \left[\sup\limits_{\w:\|\w - \w_*\| \leq \gamma_k^+}  \sum_{i=1}^n \epsilon_i\left\langle  \nabla h_i(\w) - \nabla h_i(\w_*), \w - \w_* \right\rangle \right] \\
\leq & \frac{1}{4} \E \left[\sup\limits_{\w:\|\w - \w_*\| \leq \gamma_k^+}  \sum_{i=1}^n \epsilon_i \left(p_i(\w)+q_i(\w) \right)^2 \right] + \frac{1}{4}\E \left[\sup\limits_{\w:\|\w - \w_*\| \leq \gamma_k^+}  \sum_{i=1}^n \epsilon_i \left(p_i(\w)-q_i(\w) \right)^2 \right].
\end{split}
\end{equation}

Note that the function $x^2$ is $2a$-Lipschitz over $[-a, a]$, and $p_i(\w) + q_i(\w) \in [-2\gamma_k^+ D \sqrt{\beta}, 2\gamma_k^+ D \sqrt{\beta}]$. Then, from the comparison theorem of Rademacher complexities \citep{Probability:Banach}, in particular Lemma 5 of \cite{Generalization:Bayesian}, we have
\begin{equation} \label{eqn:McD:4}
\begin{split}
& \E \left[\sup\limits_{\w:\|\w - \w_*\| \leq \gamma_k^+}  \sum_{i=1}^n \epsilon_i \left(p_i(\w)+q_i(\w) \right)^2 \right]  \\ \leq & 4 \gamma_k^+ D \sqrt{\beta}\E \left[\sup\limits_{\w:\|\w - \w_*\| \leq \gamma_k^+}  \sum_{i=1}^n \epsilon_i \left(p_i(\w)+q_i(\w) \right) \right] \\
\leq & 4 \gamma_k^+ D \sqrt{\beta} \left(\E \left[\sup\limits_{\w:\|\w - \w_*\| \leq \gamma_k^+}  \sum_{i=1}^n \epsilon_i p_i(\w) \right]+ \E \left[\sup\limits_{\w:\|\w - \w_*\| \leq \gamma_k^+}  \sum_{i=1}^n \epsilon_i q_i(\w) \right] \right).
\end{split}
\end{equation}
Similarly, we have
\begin{equation} \label{eqn:McD:5}
\begin{split}
& \E \left[\sup\limits_{\w:\|\w - \w_*\| \leq \gamma_k^+}  \sum_{i=1}^n \epsilon_i \left(p_i(\w)-q_i(\w) \right)^2 \right]\\
\leq & 4 \gamma_k^+ D \sqrt{\beta} \left(\E \left[\sup\limits_{\w:\|\w - \w_*\| \leq \gamma_k^+}  \sum_{i=1}^n \epsilon_i p_i(\w) \right]+ \E \left[\sup\limits_{\w:\|\w - \w_*\| \leq \gamma_k^+}  \sum_{i=1}^n \epsilon_i q_i(\w) \right] \right).
\end{split}
\end{equation}
Combining (\ref{eqn:McD:3}), (\ref{eqn:McD:4}), and (\ref{eqn:McD:5}), we arrive at
\begin{equation} \label{eqn:McD:6}
\begin{split}
& \E \left[\sup\limits_{\w:\|\w - \w_*\| \leq \gamma_k^+}  \sum_{i=1}^n \epsilon_i\left\langle  \nabla h_i(\w) - \nabla h_i(\w_*), \w - \w_* \right\rangle \right] \\
\leq & 2\gamma_k^+ D \sqrt{\beta} \left(\underbrace{\E \left[\sup\limits_{\w:\|\w - \w_*\| \leq \gamma_k^+}  \sum_{i=1}^n \epsilon_i p_i(\w) \right]}_{:=C_1}+ \underbrace{\E \left[\sup\limits_{\w:\|\w - \w_*\| \leq \gamma_k^+}  \sum_{i=1}^n \epsilon_i q_i(\x
) \right]}_{:=C_2} \right).
\end{split}
\end{equation}
We proceed to upper bound $C_1$ in (\ref{eqn:McD:6}). From our definition of $p_i(\w)$, we have
\[
\begin{split}
& \left| p_i(\w) - p_i(\w') \right| = \frac{1}{\sqrt{\beta}} \left| \ell'(\langle \w, \x_i\rangle, y_i)-\ell'(\langle \w', \x_i\rangle, y_i) \right| \\
\leq & \sqrt{\beta} \left|\langle \w, \x_i\rangle - \langle \w', \x_i\rangle \right| = \sqrt{\beta} \left| \langle \x_i, \w-\w_* \rangle - \langle \x_i, \w'-\w_* \rangle \right|.
\end{split}
\]
Applying the comparison theorem of Rademacher complexities again, we have
\begin{equation} \label{eqn:McD:7}
C_1 \leq \sqrt{\beta} \E \left[\sup\limits_{\w:\|\w - \w_*\| \leq \gamma_k^+}  \sum_{i=1}^n \epsilon_i \langle \x_i, \w-\w_*\rangle \right] =C_2.
\end{equation}
Next, we upper bound $C_2$ as follows:
\begin{equation} \label{eqn:McD:8}
\begin{split}
& \sqrt{\beta}  \E \left[\sup\limits_{\w:\|\w - \w_*\| \leq \gamma_k^+}  \sum_{i=1}^n \epsilon_i \langle \x_i, \w-\w_* \rangle \right] \leq  \sqrt{\beta} \E \left[\sup\limits_{\w:\|\w - \w_*\| \leq \gamma_k^+}  \left\|\sum_{i=1}^n \epsilon_i \x_i \right \| \left\| \w-\w_*\right\| \right]\\
\leq & \gamma_k^+ \sqrt{\beta} \E \left[ \left\|\sum_{i=1}^n \epsilon_i \x_i \right \|\right] \leq \gamma_k^+ \sqrt{ \E \left[ \|\x_i\|^2 + \sum_{u \neq v} \epsilon_u \epsilon_v \x_u^\top \x_v \right]} \leq \gamma_k^+ D \sqrt{\beta n}.
\end{split}
\end{equation}
We complete the proof by combining (\ref{eqn:McD:6}), (\ref{eqn:McD:7}) and (\ref{eqn:McD:8}).

\end{document}